\pdfoutput=1
\documentclass[11pt]{article}

\usepackage{acl}

\usepackage{times}
\usepackage{latexsym}

\usepackage[T1]{fontenc}
\usepackage[utf8]{inputenc}

\usepackage{microtype}
\usepackage{cleveref}
\crefname{section}{§}{§§}
\usepackage{tabularx}
\usepackage{algpseudocode}
\usepackage{algorithm}
\usepackage{graphicx}
\usepackage{comment}
\usepackage{tcolorbox}

\newcommand{\muhao}[1]{\textcolor{purple}{[\textbf{MC:} #1]}}
    \title{Take a Break in the Middle: Investigating Subgoals towards \\ Hierarchical Script Generation}
\author{
 Xinze~Li$^{1}$, Yixin~Cao$^{2\dag}$, Muhao~Chen$^{3}$, Aixin Sun$^{1\dag}$ \\ 
 \\
 $^1$ S-Lab, Nanyang Technological University \\
 $^2$ Singapore Management University
 $^3$ University of Southern California \\
\texttt{\{xinze.li, axsun\}@ntu.edu.sg} \\
\texttt{yxcao@smu.edu.sg}\hspace{1cm}
\texttt{muhaoche@usc.edu}\\
}
\begin{document}
\maketitle
\renewcommand{\thefootnote}{\fnsymbol{footnote}}
\footnotetext[2]{Co-corresponding Author.}
\renewcommand{\thefootnote}{\arabic{footnote}}
\begin{abstract}
Goal-oriented Script Generation is a new task of generating a list of steps that can fulfill the given goal. In this paper, we propose to extend the task from the perspective of cognitive theory. Instead of a simple flat structure, the steps are typically organized hierarchically --- Human often decompose a complex task into subgoals, where each subgoal can be further decomposed into steps. To establish the benchmark, we contribute a new dataset, propose several baseline methods, and set up evaluation metrics. Both automatic and human evaluation verify the high-quality of dataset, as well as the effectiveness of incorporating subgoals into hierarchical script generation. Furthermore, We also design and evaluate the model to discover subgoal, and find that it is a bit more difficult to decompose the goals than summarizing from segmented steps.

\end{abstract}

\section{Introduction}

\begin{figure}[ht]
\includegraphics[width=\linewidth]{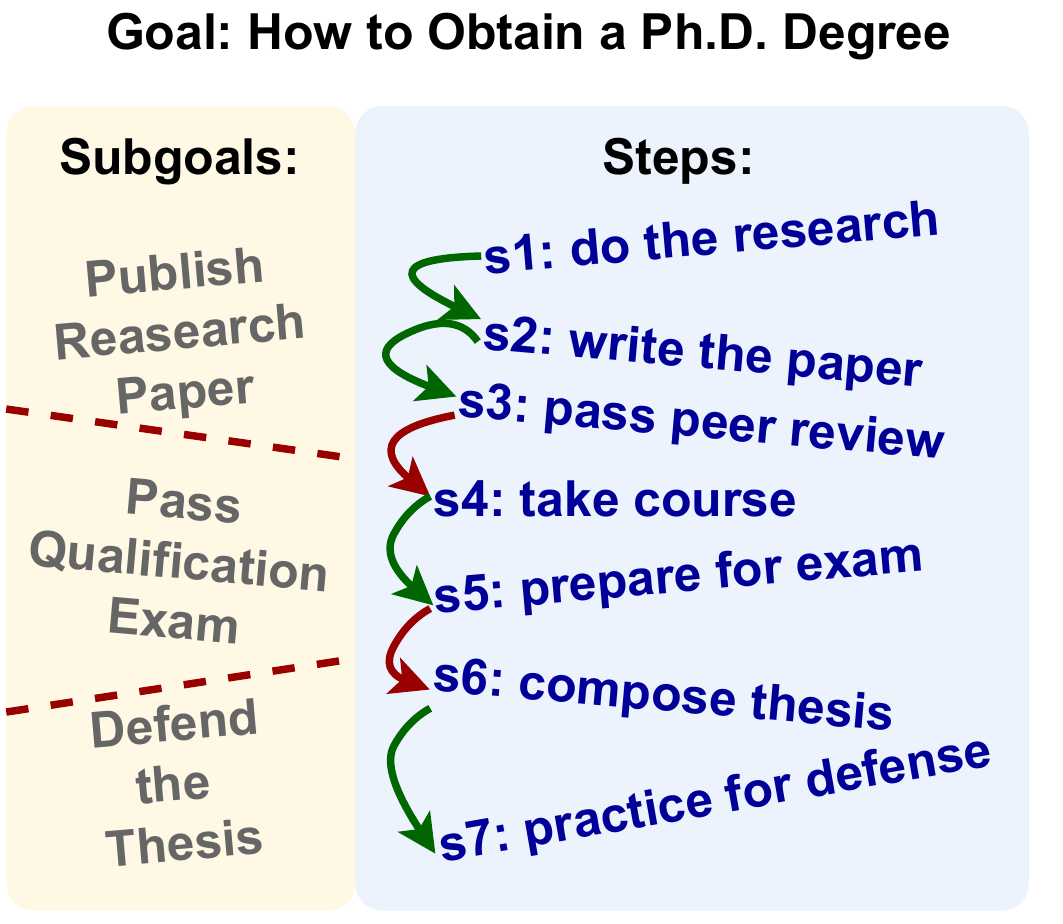}
\caption{An illustration of the hierarchical decomposition of the goal ``\textit{How to obtain a Ph.D. degree}''. We use blue for the steps, and yellow for the subgoals. Conventional task focuses on steps only, while we highlight the breaking in the middle (red arrows), which refers to the switch between higher-level subgoals.}\label{fig:goal_phd}
%\vspace{-12pt}
\end{figure}

Scripts are forms of knowledge representations for ordered events directed by particular goals~\cite{herman_1997}. As shown in Figure~\ref{fig:goal_phd}, \textit{to obtain a Ph.D degree} (i.e., goal), one shall follow specific events step-by-step, including \textit{do the research}, \textit{write the paper}, etc. Such procedure knowledge not only provides a process of problem solving, but also benefits many real-world applications, such as narrative understanding~\cite{chaturvedi-etal-2017-story}, task bots \cite{peng-etal-2021-soloist}, and diagnostic prediction \cite{zhang2020diagnostic}. Therefore, the task of script generation is proposed to automatically generate events given a goal \cite{lyu-etal-2021-goal}.

Existing works typically assume that events are sequentially arranged in a script, while we argue that this assumption leads to linear generation that is far from enough for comprehensively %acquiring script knowledge
acquiring the representation about how events are organized towards a task goal. When humans compose a script, the underlying procedure of a task is often not a simple, flat sequence. As suggested by cognitive studies \cite{botvinick2008hierarchical,zhang1994representations}, human problem solving often involves hierarchical decomposition of the task. That is to say, a more complicated task is often decomposed into subgoals, and each subgoal can be further decomposed into more fine-grained steps. For instance, the process of \textit{obtaining a Ph.D. degree} can divide into subgoals like \textit{publishing research papers}, \textit{passing the qualification exam}, and \textit{defending the thesis}. (Figure. \ref{fig:goal_phd}). The subgoal \textit{publishing research papers} thereof further consists of more fine-grained steps like \textit{doing the research}, \textit{writing the paper}, and \textit{passing peer review}. Accordingly, a proper way of acquiring script knowledge should also hierarchically capture different levels of task subgoals.

In this paper, we propose to investigate subgoals as an intermediate between goals and steps. Given a \textit{goal}, instead of generating steps in a linear manner, our task seeks to generate scripts at two levels. The first level consists of subgoals, and the second level is detailed steps, where each subgoal contains several steps. Such a new setting not only accords to people's cognitive process~\cite{antonietti2000metacognitive} but also investigates the model's abilities to understand knowledge from two aspects: problem decomposition and summarization. We propose three research questions to assist the investigation: 1) How to identify the subgoals? 2) How to effectively introduce subgoals to models to improve script generation? 3) Does the generated hierarchy align with human intuition?

%god13dog: 后面三段重点修改。方法部分少一点，dataset介绍也不用太多，更多的介绍insight。
%god13dog: in this paper, we propose to investigate subgoals 介于goal和step之间，towards heirachical script generation. 这不仅更符合人们的认知过程，也可以从两方面研究模型知识理解能力：1） 问题的decomposition （goal->subgoal, subgoal->step); 2) summerization (steps->subgoal).（也可以用原文的，不过感觉这样写更适合investigation）
%god13dog: 因此，我们要回答这几个问题：
%1） 是否存在并且可以找出subgoal？
%2）考虑subgoal是否对script generation有帮助？
%3） 怎样在模型中有效的引入subgoal和step的层次化？

%从这段开始，围绕上面的三个问题展开，可以把后面的结论和insight加以概括。对方法和数据可以简单介绍就好，比如：
%god13dog: to answer the questions, first, 我们构造了数据集，并提出多种segmentation的策略以找出subgoal。数据集xxx，segmentation+subgoal labeling xxx。基于此，我们定量+定性的探究了subgoal的质量。
%god13dog: 另起一段，second（对应第二个问题）， 我们构建了benchmark with several baselines。xxx。我们观察到，通过引入subgoal（无论自动的还是gold），在steps的流畅度、多样性、和实现目标（？）都有了一定的提升。
%god13dog: 第三段，third，我们进行了大量的实验。1）如fig2的insight，不考虑subgoal的baseline在更多的segmentation script上，效果是下降了，通过考虑预测的，得到提升。然而，距离gold还有很大差距。（不用都写，概括就行，比如在不同的segmentation上效果不同，越长越差，但是引入gold会提升）2）human eval，i）decompose xxx ii）subgoal summerize step xxx，哪个更难？ 3）error ana 。

To answer the first question, we construct a new dataset, namely ``Instructables''\footnote{Our code and dataset are available at:\\\url{https://github.com/lixinze777/Hierarchical-Script-Generation}}, about D.I.Y projects involving goals, subgoals, and steps. Besides, we extend the existing wikiHow dataset~\cite{zhang-etal-2020-reasoning} with the subgoals. To automatically obtain subgoals, we deploy a segmentation method that separates steps in training data. For each segment, we further leverage a prompt-based fine-tuning~\cite{lester-etal-2021-power} method to learn subgoal generation. We have designed quantitative metrics for evaluation, which verifies the reasonability of the two-level hierarchy of script. 

For the second question, we build the benchmark with multiple baselines. The basic idea is to incorporate the goal, subgoals, and steps into one prompt template, and use special tokens to preserve structure information. Then, we finetune Pre-trained Language Models (PLMs) to generate in a top-down or interleaving manner. This allows the model to \emph{take a break} to conclude on each subgoal before generating the succeeding steps. We have conducted extensive experiments. The results show that by including subgoals, the model generates scripts with better soundness, diversity, and making better sense to achieve the goal. In fact, given gold standard segmentation and subgoals, the improvement is more substantial, indicating space for improvement in our predicted subgoals.

To address the third question, we conduct human evaluation to assess the quality of both steps and subgoals, as well as the above two types of model abilities. We observe that the language model shows a weaker ability to decompose the goals than to summarize the steps. We have also analyzed the errors in detail and found that the models sometimes generate repetitive subgoals and low-quality steps. The models still have much room for improvement in generating high-quality hierarchical scripts if the goal is too complicated.

To summarize, our work has three contributions:
1) We construct the dataset ``Instructables'' and extend the wikiHow dataset to investigate the subgoals as intermediate between goals and steps.
2) We build a benchmark with several baselines towards hierarchical script generation.
3) We conduct extensive evaluation to investigate the effects of subgoals qualitatively and quantitatively.

%summarize contribution 

%1 investigate subgoal + dataset
%2 method + benchmark
%3 experiment 

%subgoal 介于goal step之间 - 》 两层
%. task: hierarchical generation with subgoal + dataset
% benchmark
\section{Related Work}
\paragraph{Procedural Knowledge Acquisition}
Early research on the script and procedural knowledge is usually formulated as a classification or ranking problem. For example,  \citet{modi-titov-2014-inducing} and~\citet{ pichotta-mooney-2016-using} predict a score for each given event to determine their relative order based on event representation learning.
%More works started to predict relevance and ordering \cite{zhou-etal-2019-learning-household, zhang-etal-2020-reasoning} leveraging the wikiHow dataset.
P2GT~\cite{chen-etal-2020-trying} and APSI~\cite{zhang2020analogous} further analyze the intention of events and conduct a joint learning-to-rank for better ordering. Thanks to the success of the Pre-trained Language Model (PLM), recent work GOSC~\cite{lyu-etal-2021-goal} proposes to generate steps in a simple, flat manner for any goal. Another line of works \cite{pareti2014integrating, lagos2017enriching} have attempted to establish a hierarchical structure among scripts by linking their steps. Given any event of goal \textit{A}, \citet{zhou-etal-2022-show} compute a similarity score to find the most semantically close goal \textit{B}, so that all events of  \textit{B} can be regarded as detailed subevents of the given event at the lower level. This approach, although effective, has an exceptionally high demand on the dataset to cover a wide range of goals. In many cases, there is no reasonable goal for alignment, which results in a deviation in the meanings between the linked sentences. Therefore, we neither regard steps in a flat manner, nor link steps and goals with the retrieve-then-rerank approach. Instead, we propose a task and model targeting the inner hierarchy of a script during generation and are complementary to the above works.

\paragraph{Controlled NLG}
Script generation is a form of controlled text generation task~\cite{hu2017toward} since the generated scripts are attributed to the given goal. To increase the controllability of text generation, research efforts investigate the ways of constrained decoding. NeuroLogic Decoding~\cite{lu-etal-2021-neurologic} improves controlled generation upon semantic constraints by enforcing predicate logic formula at the decoding stage. NeuroLogic A*esque Decoding~\cite{lu-etal-2022-neurologic} further incorporates a lookahead heuristic to estimate future constraint satisfaction. Controlled text generation tasks can take other forms like generating descriptions conditioned on subparts of a table~\cite{wang2022robust}. Another classic application of controlled text generation is storytelling, whereby stories are generated based on a prompt or a storyline. \cite{fan-etal-2018-hierarchical} generate hierarchical stories conditioned on a prompt that was generated first. \cite{fan-etal-2019-strategies} enhance the coherence among different levels of a hierarchical story with a verb-attention mechanism. Unlike tasks like storytelling, script generation is not open-ended since it is goal-oriented.  
\section{Task and Dataset}
In this section, we first formulate the new task setting and then introduce the new dataset that we constructed, namely Instructables.

\subsection{Task Definition} \label{def}
%\textbf{Conventional 
The original
\textit{goal-oriented script generation
}~\cite{lyu-etal-2021-goal} (GOSC) focuses on generating a \emph{sequence} of steps (or events) that accomplishes the given goal. In contrast, the proposed \textit{hierarchical script generation} conducts hierarchical generation and models the script as multiple levels of events.
Formally, given a \textit{goal g} as input, it is to generate $L$ levels of events as output, where the events at the $1$-st to the $(L-1)$-th levels are called \textit{subgoals (s)} and the events at the $L$-th (most fine-grained) level are called \textit{steps (t)} . Within each level, the list of children events should fulfill the objective of their parent event. Note that the number of events at each level is not fixed, and the model is required to decide the number by itself. Based on our observation, two levels of events are sufficient for most scripts (i.e., $L=2$) in reality. For example, both websites, wikiHow and Instructables, define several sections for each goal, and each section contains multiple steps. 
These task instruction resources are all organized in two levels.
Thus, in the rest of the paper, we define $L=2$.

This task inherently include 2 settings. The input for both settings are the goal g. Setting 1 takes the sequence of events from the lowest level of the hierarchy as output, which is the same as the GOSC task. Through this setting, we investigate whether including subgoals improve the traditional script generation task. Setting 2 takes the entire hierarchy of events as output. Through this new setting, we investigate the language model's ability to generate high-quality subgoals and steps.
% task definition风险 - 2层

\subsection{Dataset} \label{dataset}
We use two datasets, wikiHow and Instructables, for the studied task. The wikiHow dataset \cite{zhang-etal-2020-reasoning} is a collection of how-to articles crawled from the wikiHow website. Each article describes methods to achieve a given goal in the title. The articles are written in sections, each of which comes with a few steps. In this work, we consider one section of steps as one segment and a section name as a subgoal. Due to the lack of resources, many research works on script use wikiHow as the only dataset.

To verify the model's generalizability over more than a single dataset, we construct another dataset based on Instructables\footnote{\url{https://www.instructables.com/}} --- a website specializing in user-created do-it-yourself (DIY) projects.

%\paragraph{
\noindent \textbf{Instructables Construction}
The data construction process consists of two stages. The first is raw data preparation. We collect the content of each project according to its category (Circuits, Workshop, Craft, Cooking, Living, and Outside) using Scrapy.\footnote{\url{https://github.com/scrapy/scrapy}} Each project consists of a title showing the item that the author sought to make (i.e., a toy rocket) and the instructions to make this item. In most cases, authors write the instruction in a few sections, each with a step-by-step description. During crawling, We take each section name as a subgoal and every sentence as a step. 

The second stage is filtering. The raw data is inconsistent in text style, section format, article length, etc. We hereby carry out seven steps to filter out noisy data. We remove: 1) Non-English projects using Python library langdetect.\footnote{\url{https://pypi.org/project/langdetect/}} 2) The section on Supplies or Materials, which describes composed materials instead of events/actions. 3) The section numbers (e.g., \textit{``section 3: draw a line''} -> \textit{``draw a line''}). 4) Unnecessary spaces and characters like Line Feeder to maintain the human-readable text. 5) Projects with empty text content in any section since %they normally use
these sections are usually presented as figures or videos. 6) Projects with any overly lengthy section. Many authors post stories or anecdotes to convey the rationale they came up with the project, and we find 128 words a good threshold to filter them out. 7) Projects that build the same item as others. We remove repeated articles about seen items to eliminate possible redundancy or anomalies in data distribution. Finally, we unify the format of project titles to make them consistent. By performing Part-of-Speech (POS) Tagging on titles, we prefix ``How to make'' to noun phrases (e.g., \textit{How to make} Kung Pao Tofu), we recover the verb from stemming and prefix ``How to'' to verb phrases (e.g., \textit{How to} Build a Toy Rocket), and we retain the How-to questions (e.g., How to Create a Puppet).

\begin{table}[t]
\centering
{
\begin{tabular}{l|ccc}
\hline
\textbf{Category} & \textbf{Scripts} & \textbf{Subgoals} & \textbf{Steps}\\
\hline
\textbf{Circuits} & 22,437 & 109,917 & 282,685\\
\textbf{Workshop} & 16,991 & 94,554 & 257,248\\ 
\textbf{Craft}   & 24,874 & 137,471 & 365,244\\ 
\textbf{Cooking}  & 12,916 & 69,633 & 189,371\\
\textbf{Living}  & 23,204  & 114,113 & 291,682\\ 
\textbf{Outside}  & 6,986  & 34,391 & 92,439\\\hline
\textbf{TOTAL}  & 107,408  & 560,079 & 1,478,669\\\hline
\end{tabular}
}
\caption{Number of total scripts, subgoals, and steps in dataset Instructables by category.}
\label{tab:dataset statistics}
%\vspace{-12pt}
\end{table}

\paragraph{Dataset Statistics}
Table~\ref{tab:dataset statistics} shows the statistics of Instructables by category. In total, we obtain 107,408 scripts, 560,079 subgoals, 1,478,669 steps, and 26,813,397 words. We analyze the difference between Instructables and wikiHow in \cref{app: compare}

%statistic比对可以放在附录
\begin{comment}
\begin{figure*}[ht]
\includegraphics[width=\textwidth]{figures/flowchart2.pdf}
\caption{The overview of our proposed method. In the first module, flat training data is separated into different sections through segmentation. In the second module, a T5-base predictor is trained to predict he subgoal of each section obtained from the previous module. In the third module, the steps are combined with their subgoals to form a template, prompted into a T5-base model for fine-tuning and generation. }\label{fig:flowchart}
\end{figure*}
\end{comment}

\section{Method}
In this section, we present our series of methods to build the benchmark for the hierarchical script generation task. Note that we do not target a best-performing model but aim at a reasonable baseline to shed light on future research. We first introduce a segmentation method that automatically segment the steps and generate their subgoals, in case there is no ground truth hierarchy available for training. Then, given the goal, subgoals, and steps, we introduce the proposed framework for training. Finally, given a goal, we detail the inference process.  
%The proposed method conducts learning in a multi-modular framework as illustrated in the flowchart (Fig. ). Pretrained language model is first used to produce the sentence embedding for each step. Each script in training dataset is segmented into likely sections by grouping consecutive steps that can achieve a subgoal according to the steps' embeddings. Then, we fine-tune a pretrained model on goal prediction task by feeding in the steps as input and the goal as output. Finally, we join the segmented steps with its predicted section goal and conduct a prompt-based learning. We fine-tune a language model to generate hierarchical scripts with subgoal given a goal. 

%In the rest of this section, we introduce the technical details of each step for segmentation, subgoal prediction and prompt-based learning.
\subsection{Segmentation}
The dataset wikiHow and Instructables naturally manifest an easy-to-consume hierarchical layout, as we explained before. However, for better generalization, we do not assume that all sources of script data possess the privilege of a hierarchical layout with sections and subgoals. Formally, given an ordered list of steps, we seek to find segmentation points between steps to separate them into relatively concrete segments. Each segment should inherently represent a subgoal. To find these segmentation points in an unsupervised manner, we propose four methods. The first method finds low probability scores between consecutive steps via BERT \textbf{next sentence prediction}~\cite{devlin-etal-2019-bert}. The second method measures the plausibility of a list of steps with \textbf{perplexity} and locates the abnormal ones. The third method applies \textbf{clustering} algorithm to group steps based on their sentence embeddings. The last method locates segmentation points upon multiple \textbf{topics detected} via fastclustering~\cite{reimers-2019-sentence-bert}. We explain these methods in detail in \cref{app: segmentation}.

\subsection{Subgoal Labeling}
Given steps and their segmentation, we are to generate an event for each segment as their parent subgoal, where the subgoal is a high-level summarization of its children steps. Due to the lack of annotations, we perform the labeling in a self-supervised manner. That is to say, we regard it as a dual problem of script generation. Given a goal and all the steps in a flat format, instead of training a model to generate steps, we fine-tune a T5-base model \cite{raffel2020exploring} to generate the goal using the list of steps as inputs. Specifically, We convert the question-format goal into a verb phrase by removing the ``How to'' prefix, which is more suitable as a subgoal. Note that we did not include any additional training data but reused the training dataset for the script generation task. This practice ensures that the system is not leaked with any sentences in the development or testing data to gain unfair information at training time.

\subsection{Hierarchical Generation} \label{prompt}
\paragraph{Training}
Given a goal, we train a model to generate a varying number of subgoals, and each subgoal has its own steps. Thanks to the recent progress of prompt-based fine-tuning~\cite{lester-etal-2021-power}, we use the special token \texttt{<section>} to preserve the structure information~\cite{yao2019kg} and take advantage of PLMs to generate such a two-level structure by re-formatting a prompt: 

%\muhao{could use a greybox.}

\begin{tcolorbox}
\small
{[Goal]}, \texttt{<section>} With {[Subgoal]}, {[steps]}. \texttt{<section>} With {[Subgoal]}, {[steps]}
\end{tcolorbox}

An example prompt for the goal \textit{How to learn Web Design} is as below:

\begin{tcolorbox}
\small
\textit{To learn Web Design, \texttt{<section>} with Finding Web Design Resources, check online for web design courses and tutorials. Look into taking a class at a local college or university... \texttt{<section>} With Mastering HTML, familiarize yourself with basic HTML tags. Learn to use tag attributes... }
\end{tcolorbox}

Intuitively, there are two typical generation sequences for a multi-granular generation task, \textbf{interleaving} and \textbf{top-down}. The above template adopts an interleaving generation sequence. In terms of the auto-regressive decoding sequence, a subgoal is generated, followed by its affiliated steps, the next subgoal, and so on. We also propose a template incorporating a top-down generation sequence where all subgoals are generated first followed by all the steps. The prompt is as follows: 

\begin{tcolorbox}
\small
{[Goal]}, the subgoals are: {[Subgoal]}, {[Subgoal]}. \texttt{<section>}, {[steps]}. \texttt{<section>}, {[steps]}
\end{tcolorbox}

For both generation sequences, we add a special token \texttt{<section>} as a delimiter between two segments to denote the hierarchy information. Of course, for more levels, one can add more special tokens as delimiters between different levels. Inspired by~\citet{yao2019kg},  we use a special token to delimit different parts of the structure instead of conducting a complex hierarchy decoding process. This technique leads to two benefits. First, it allows a sequential generation process that aligns well with the pre-training objective of PLM, therefore facilitating knowledge prompting from the PLM. Second, the hierarchy information improves long-text generation (e.g., sometimes there are many steps to decode) because the subgoal shortens the dependency between steps by providing a high-level summarization. We leave the exploration for other long-text generation tasks in the future.   

\paragraph{Inference}
At inference time, we feed the how-to question, which contains the goal, into the tuned PLM as input, with the prefix ``Ask question:'' as common practice for Question Answering tasks using the PLM. We fix the hyper-parameters the same as across training settings. The decoder generates subgoals and steps in an interleaving/top-down approach, and the output is the same as the prompt format we design for training sets. We leverage the special tokens in output as the beacon for extracting subgoals and subsequently expand the linear output into the hierarchical format.
\begin{table*}[t]
  \centering
  {
  \begin{tabular}{l|l||ccccc}\hline
    \textbf{Dataset} & \textbf{Method} & \textbf{Bert. $\uparrow$} & \textbf{Perp. $\downarrow$} & \textbf{BLEU-1$\uparrow$} & \textbf{ROUGE-L $\uparrow$} & \textbf{Dist.-3 $\uparrow$} \\\hline
     & GOSC (mT5-base) & 82.3 & 17.0 & - & - & -\\
     & GOSC & 85.3 & 14.3 & 20.4 & 20.2 & 93.4 \\
    \textbf{wikiHow}  & GOSC (two-stage) & 83.6 & 20.9 & 16.2 & 22.6 & 93.7 \\
    & HSG (top-down) & 85.5 & \textbf{12.8} & 22.9 & 23.2 & \textbf{94.0} \\
     & HSG (interleaving) & \textbf{85.6} & 15.4 & \textbf{23.7} & \textbf{24.1} & \textbf{94.0}\\\cline{2-7}
    & HSG w gold subgoal & 85.8 & 12.7 & 31.2 & 27.0 & 95.9 \\\hline
     & GOSC  & 82.2 & 31.3 & 16.6 & 16.0 & 92.5 \\
    \textbf{Instructables} & HSG (top-down) & 82.0 & \textbf{19.5} & 17.4 & 17.8 & 94.4 \\
     & HSG (interleaving)  & \textbf{82.8} & 22.2 & \textbf{18.9} & \textbf{21.8} & \textbf{94.5} \\\cline{2-7}
     & HSG w gold subgoal  & 82.8 & 24.3 & 25.1 & 23.4 & 96.5 \\\hline
  \end{tabular}
  }
  \caption{Performance of our method (marked as HSG) in top-down and interleaving approach on test set of wikiHow and Instructables datasets. We also report the cases where gold segments and subgoals used in the generation process as a performance upper bound. We abbreviate BERTScore, Perplexity, and Distinct-3 to Bert., Perp., and Dist.-3 respectively. We \textbf{bold} the best performance. }
  \label{tab:main_result}
\end{table*}

\section{Experiments}
To evaluate the proposed framework for hierarchical script generation, we conduct extensive experiments on the presented wikiHow and Instructables datasets. We compare our proposed framework with prior strong baseline method and discuss the results (\cref{Automatic Evaluation}-\cref{Human Evaluation}). We have also investigated the best segmentation method as a secondary experiment. (\cref{Segmentation}).We conduct ablation studies (\cref{ablation study}).
% bert 83.6 perp 20.9 bleu 16.2 rouge 22.6 dist 94.7
\subsection{Experimental Setup} \label{Experiment Settings}
\paragraph{Dataset}
Following the setup from \citet{lyu-etal-2021-goal}, we randomly separate datasets into training and test sets using a 90/10\% split, and hold out 5\% of the training set as the development set. We perform both automatic and human evaluations to assess the quality of the generated script.

\paragraph{Metrics}
For automatic evaluation metrics, we first follow prior works~\cite{lyu-etal-2021-goal} without considering the hierarchy information as our task setting 1 \cref{def} and report \textbf{perplexity} and \textbf{BERTScore} \cite{zhang2019bertscore} for steps. Perplexity is the exponential of log-likelihood of the sequence of steps assigned by the model. In addition, we compute three widely used generation metrics, \textbf{BLEU-1} \cite{papineni-etal-2002-bleu}, \textbf{ROUGE-L} \cite{lin-2004-rouge} and \textbf{Distinct-n metric} \cite{li-etal-2016-diversity} by taking average over all testing data. For our experiment that investigates the best segmentation strategy, due to the lack of measurements, we define a metric ``segment distance'' and the details can be found in \cref{Segmentation}. When evaluating the hierarchical scripts, we remove the subgoals and prompt template to ensure fair comparison.

\paragraph{Baseline}
Given the different nature of the hierarchical script generation in contrast to the previous task, there is not a directly applicable baseline. Here, we choose a state-of-the-art model for conventional script generation, namely GOSC~\cite{lyu-etal-2021-goal}, as a strong baseline. Note that we carefully re-implement the model according to the settings and parameters from \cite{lyu-etal-2021-goal} and report a variance where the mT5-base model of GOSC is replaced as T5-base model~\cite{raffel2020exploring} to learn better on the English corpus, since GOSC was originally designed for a multilingual setting. Another baseline is a two-stage generation process. First generating the steps in GOSC manner, then using these steps as subgoals to generate the actual steps.

\subsection{Automatic Evaluation} \label{Automatic Evaluation}

We report the average results of 3 runs of hierarchical script generation on both datasets in Table \ref{tab:main_result}. We compare the quality of the generated text according to the four metrics (\cref{Experiment Settings}). For each dataset, we also report the case where ground truth segmentation and subgoals (section name) are provided as a performance upper bound (of the proposed prompt-based fine-tuning method, as better performance may be achieved by more advanced model).

We can observe that the results from both datasets are generally consistent. \textbf{(1)} The two-stage baseline is weaker than the basic baseline as indicated by most metrics. This baseline has an error aggregation problem whereby the steps generated could be irrelevant after two stages of generation. \textbf{(2)} Our method has outperformed the baseline in three metrics (Perplexity, BLEU-1, and ROUGE-L scores), indicating the effectiveness of our method in generating scripts with higher quality. \textbf{(3)} The improvement in Distinct-3 metric on both datasets indicates that our method is capable of generating texts of greater variation. With segmentation, we take a break in the middle of the generation process, and provides the model a chance to conclude on steps of the current subgoal, and refer to the information from the upper level of the hierarchy. The model thereafter generates the script with better quality and less repetition. \textbf{(4)} Between the top-down and interleaving approaches, the latter achieves slightly better or tied scores among almost all metrics for both datasets. The subgoals are in proximity to the corresponding steps for the interleaving approach, which better guides the step generation. \textbf{(5)} It is noteworthy that our method with predicted subgoals outperforms the gold segmentation and subgoals on perplexity for Instructables dataset, showing that our generated subgoals might be closer to natural language compared to using gold subgoals from Instructables. \textbf{(6)} Except for this, using gold segmentation and subgoal leads to better results in all other metrics on both datasets, manifesting an enormous potential of our method, which also indicates an area of improvement for the accuracy of our predicted segmentation and subgoals. 
We acknowledge that existing metrics are unable to directly measure how well a script fulfills a goal. Hence, we conduct human evaluations to complement automatic evaluation.

\begin{comment}
\paragraph{Subgoals}
\begin{table}[ht]
  \centering
  \begin{tabular}{l|ccc}\hline
    \textbf{Dataset} & \textbf{Bert.} & \textbf{B-1} & \textbf{R-1} \\\hline
    \textbf{wikiHow} & 90.7 & 26.4 & 38.1 \\
    \textbf{Instructables} & 88.5 & 20.1 & 35.3 \\\hline
  \end{tabular}
  \caption{subgoal  }
  \label{tab:autoeval_hierarchy}
\end{table}
\end{comment}
\begin{figure}[t]
\includegraphics[width=\linewidth]{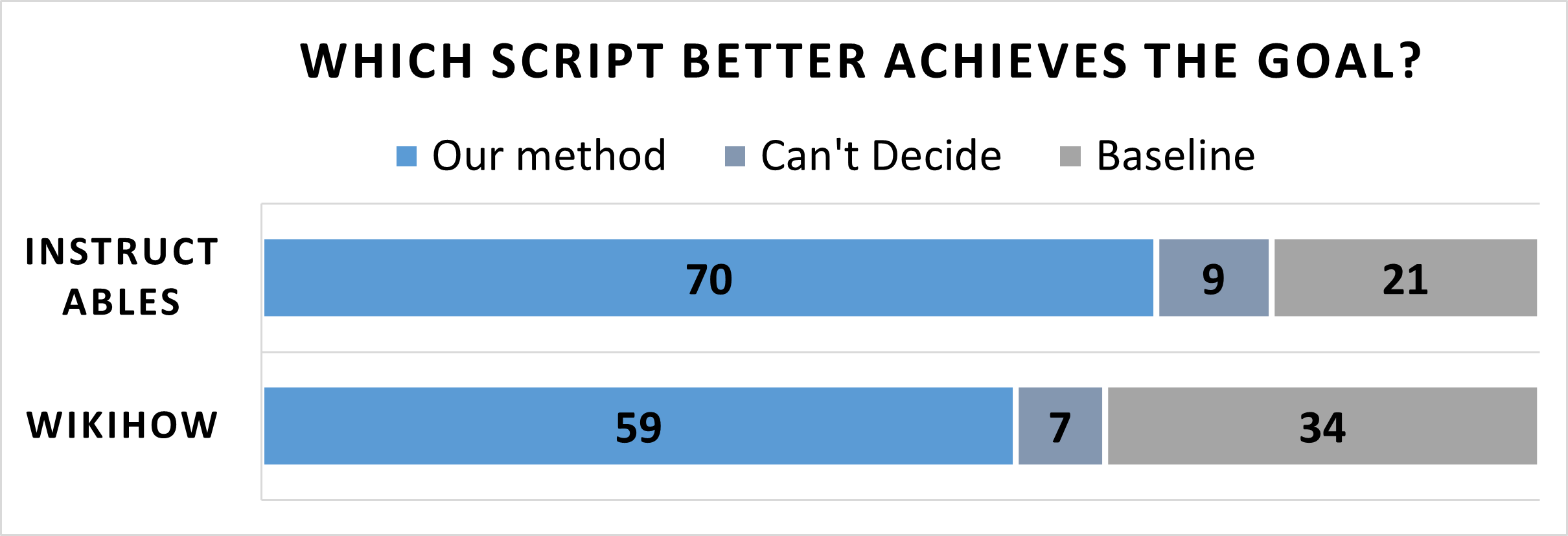}
\caption{Stacked bar chart showing the result of human evaluation on question related to steps. Blue, grey and orange color respectively indicates the percentage that our script is preferred, baseline script is preferred, and cannot decide.}\label{fig:human_eval_steps}
\end{figure}
\subsection{Human Evaluation} \label{Human Evaluation}
We conduct human evaluations to assess the quality of our hierarchical scripts for task setting 2 \cref{def}. We evaluate the scripts on both the steps and the subgoals. The steps are assessed through direct comparison, where we ask the annotators to choose between scripts generated by our method (flattened) and by the baseline. In addition, we also evaluate the generated subgoals based on two criteria. The first criterion concerns whether the annotators consider the generated subgoals as valid components of the task goal, i.e., problem decomposition ability. In the context of the main goal, the second criterion concerns if the generated subgoal properly represents the associated steps, i.e., summarization ability. We provide more details (e.g., guideline and annotators) of human evaluation in \cref{app: human evaluation questions}.

\paragraph{Step Evaluation}

From Figure~\ref{fig:human_eval_steps}, the scripts generated by our method were more favored by annotators over the baseline scripts in both wikiHow (59\%) and Instructables (70\%) test sets than the baseline scripts, excluding 7\% and 9\% draw cases. In addition, we realize that the proportion of favored scripts is higher on the Instructables dataset than on wikiHow. Such results are due to the high-quality wikiHow scripts generated by the baseline method. Our method has a greater improvement on Instructables, similarly, attributes to the low-quality scripts from the baseline, mainly because Instructables' scripts are more difficult and longer, which requires the ability of problem decomposition. In extreme cases, we observed empty or very short outputs for Instructables using the baseline method, which did not appear in the scripts generated by our method. Further, we analyze more typical examples and mistakes in case study (\cref{case study}).

\paragraph{Subgoal Evaluation}
\begin{figure}[t]
\includegraphics[width=\linewidth]{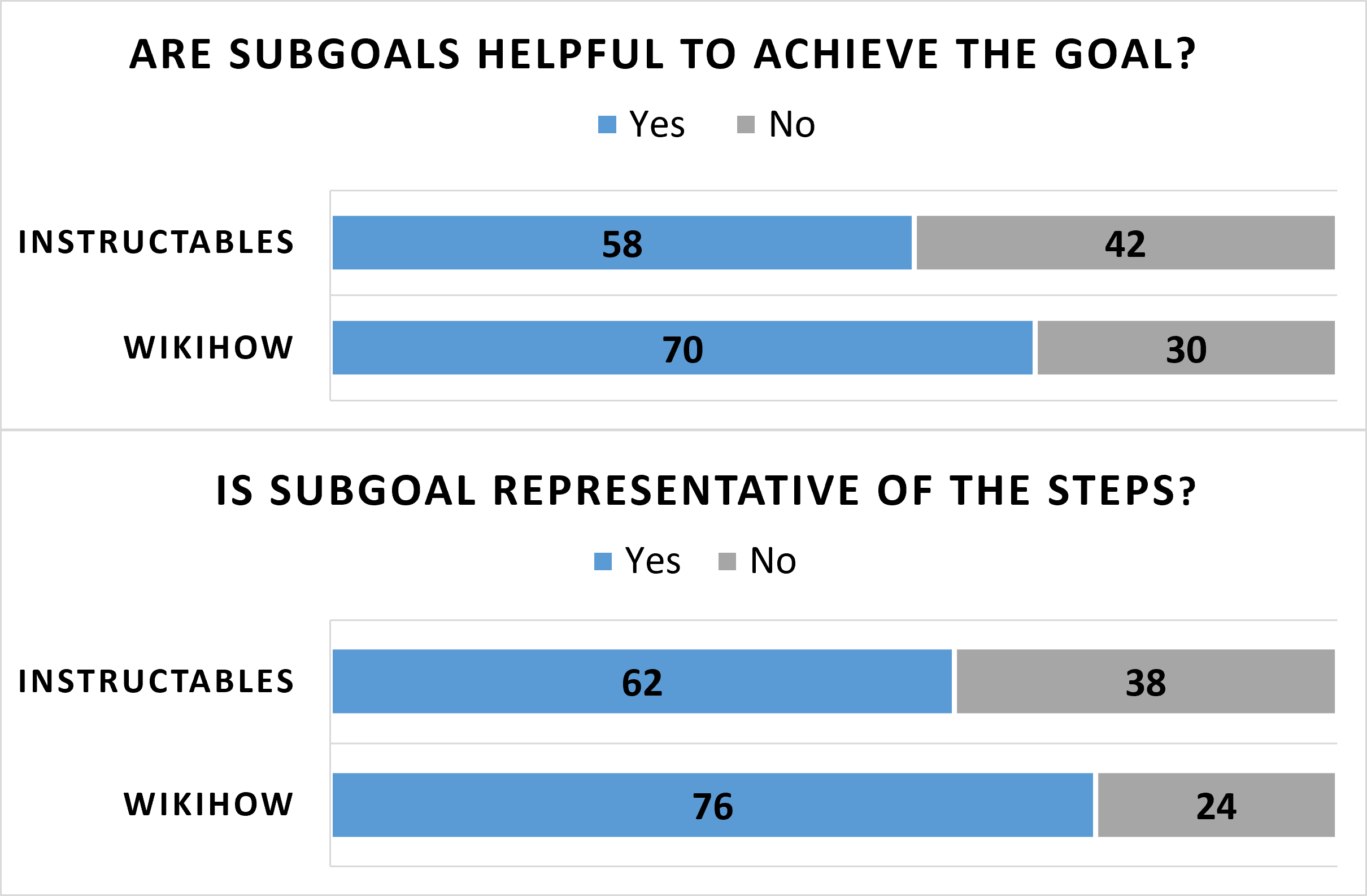}
\caption{Results on subgoals. Blue and grey color respectively indicates the percentage of ``Yes'' and ``No''}\label{fig:human_eval_subgoals}
\end{figure}
From Figure~\ref{fig:human_eval_subgoals}, regarding the question of whether subgoals are helpful to achieve the goal, 70\% of the subgoals are given credit by the annotators for the wikiHow dataset, while this percentage is 58\% for the Instructables dataset. For the other question assessing whether generated subgoal well-represents the associated steps, the percentage of positive responses for the wikiHow dataset (76\%) also surpasses that of the Instructables dataset (62\%). The results from these two questions accord with each other that the subgoals generated for the Instructables dataset are of worse quality than that of wikiHow. From another perspective, comparing the results between two questions, we find that the generated subgoals have a weaker degree of association with the goals than with the generated steps. The results demonstrate a great challenge on complex task decomposition.

\subsection{Segmentation} \label{Segmentation}

\begin{table}[t]
\centering
{
\begin{tabular}{l||cc}
\hline
\textbf{Segmentation Method} & \textbf{$k$=$3$} & \textbf{$k$=$4$}\\
\hline
\textbf{2 Segments} & 4.17 & 5.20\\
\textbf{3 Segments} & 4.23 & 5.16\\\hline 
\textbf{Next Sentence Prediction} & 4.23 & 5.26\\ 
\textbf{Perplexity}  & 4.30  & 5.36\\
\textbf{Agglomerative Clustering}  & 4.08  & 4.89\\
\textbf{Topic Detecting}  & \textbf{3.82} & \textbf{4.69}\\\hline
\end{tabular}
}
\caption{Segment distances of the proposed segmentation methods on wikiHow dataset.}%\muhao{why is the font different in this table?}}
\label{tab:segmentation result}
%\vspace{-12pt}
\end{table}

To better understand the best segmentation strategy for our task, we assess distinct segmentation techniques and aim to find the one with the closest segment structure to the ground truth. In order to measure the affinity between predicted and gold segmentation points, we propose the metric ``segment distance'' in light of the metric ``edit distance'' in quantifying string similarity. Instead of calculating the number of minimal editing scripts, ``segment distance'' calculates the least number of steps to shift $m$ predicted segmentation points to the actual ones, where $m$ = min($p-1$, $g-1$), $p$ is the number of predicted segments, and $g$ is the number of segments in ground truth. In addition, we impose of penalty score $P$ = $k*d$ for difference of number of segments $d$ = |$p-g$| as a metric that encourages accurate estimation of number of segments. The $k$ value is set between 3 and 4 as a fair penalization.
% p: number of predicted segment
% p-1: number of predicted segmentation point
% g: number of gold segment
% g-1: number of gold segmentation point
% m = number of points to shift ,m = min(p-1, g-1)

% select m points from predicted or gold (depend on which one is fewer) such that S is minimized
% let the predicted point be p1, p2 .. pm
% let the gold points be g1, g2, .. gm
% S: Number of Shift, S = summation(|pi-gi|) [i = 1 to m]

% d: difference in segmentation point d = |(p-1)-(g-1)| = |p-g|
% k: penalty coefficient, k = 3 or 4 in practice
% P: Penalty Score, P = k*d

% Segment Distance = Number of Shift (S) + Penalty Score (P)

As a baseline, we take the average number of segments $N$ (closet integer) from the dataset and carry out an $N$-equal splits on each script. This simple approach is in fact a strong baseline since most scripts have a limited number of steps in different segments. This baseline inherently maintains a small penalty score $P$. We evaluate the segmentation performance on 1,000 random scripts.

In Table \ref{tab:segmentation result}, we report the results of segmentation experiment on wikiHow dataset as a representation, where a smaller average segment distance indicates a structurally similar segmentation to the gold standard. We report two settings on $k$ = $3$ and $4$ respectively. On the one hand, the baseline method is demonstrated to be strong since they produce comparable results with NSP and Perplexity methods. On the other hand, the Clustering and Topic Detecting methods outperform the baselines. Topic Detecting produces the best score of $3.82$ when $P$ = $3$, and $4.69$ when $P$ = $4$. As such, we chose it as our segmentation method for this work.

\subsection{Ablation Studies}\label{ablation study}

\paragraph{Scripts: Long vs. Short}
\begin{figure}[t]
\includegraphics[width=\linewidth]{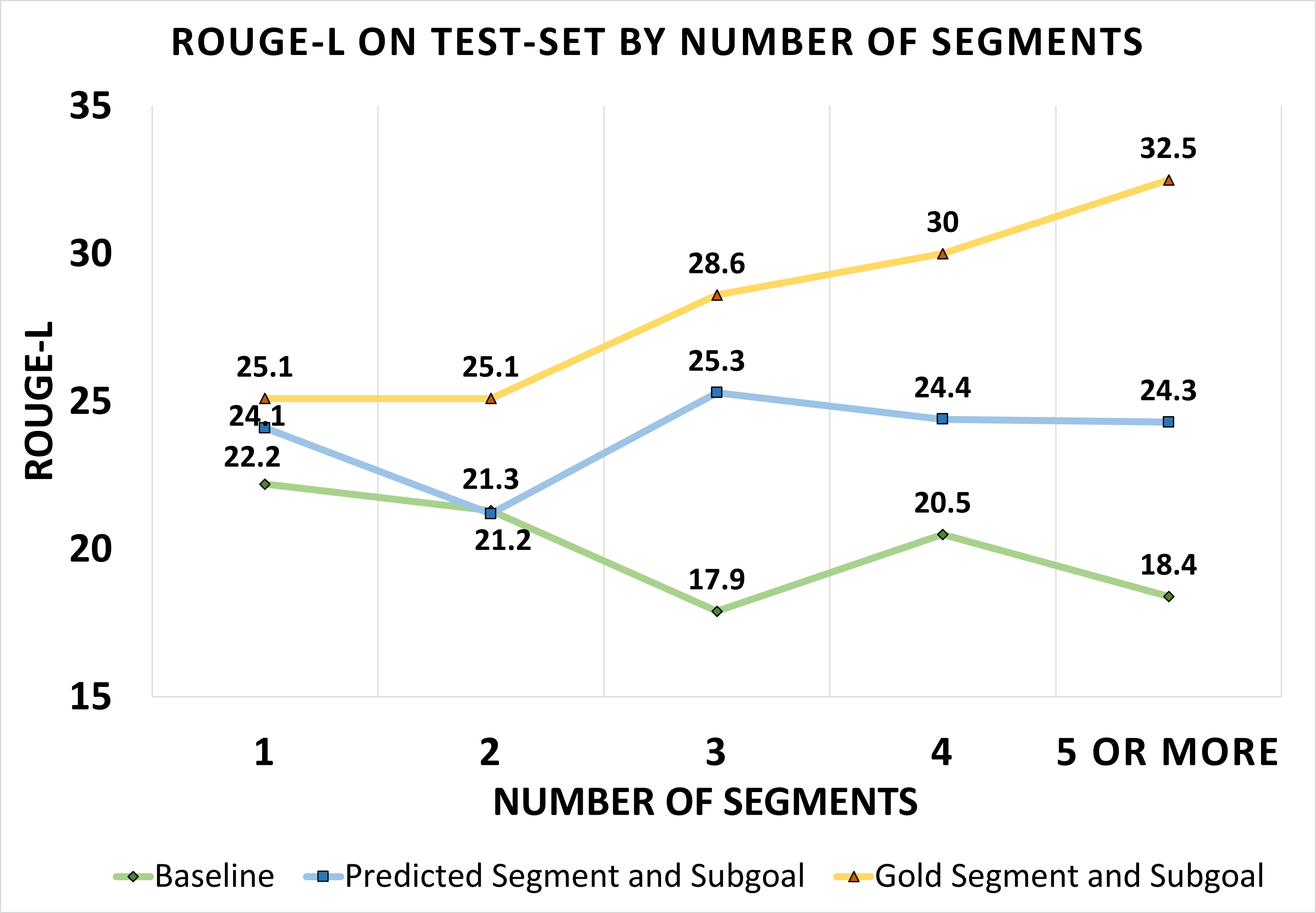}
\caption{ROUGE-L Score on test set categorize by number of segments from ``1'' to ``5 or more''. Green color represents the baseline method. Blue color represents our method. Yellow color represents our method using gold segmentation and subgoals.}\label{fig:goal_ablation1}
%\vspace{-12pt}
\end{figure}
On wikiHow, about 25\% of the scripts are written in one whole segment without subgoals. We are interested to see whether our method can improve the generation of these scripts as well. We extend this question by navigating the impact of our solution on scripts of different lengths and with different number of subgoals. The dividing of the segments can vary according to the complexity of the goal, and the author's writing style --- some authors prefer to write scripts all in one segment. We categorize the test set according to number of segments in ground truth scripts into ``1'', ``2'', ``3'', ``4'' and ``5 or above'', where the number of steps are averaged as 9.5, 13.3, 16.1, 20.8 and 27.9, respectively. We select ROUGE-L as a representative metric, and plot a graph of performance with respect to the number of segments in Figure \ref{fig:goal_ablation1}. From the result, %\textbf{(1)} 
it is evident that the improvement is outstanding with long scripts (more segments) in test set. %\textbf{(2)} 
Overall, the baseline scripts showcase a downward trend as the number of segments increases, since decoding is often more challenging when the texts are longer -- this is a common difficulty for long text generation. Our methods tackle this problem by taking a break in the middle and providing room for adjustment at decoding time with segmentation and subgoals. %\textbf{(3)} 
Consequently, hierarchical generation manifests a rising trend, especially with gold segment and subgoals. %\textbf{(4)} 
Another interesting phenomenon is that the performance improves, although not significantly, for the single segment scripts. Although the authors do not divide segments when composing scripts, the tasks may still inherently manifest a hierarchical structure, as solving a task naturally falls into different stages.
\begin{figure}[t]
\includegraphics[width=\linewidth]{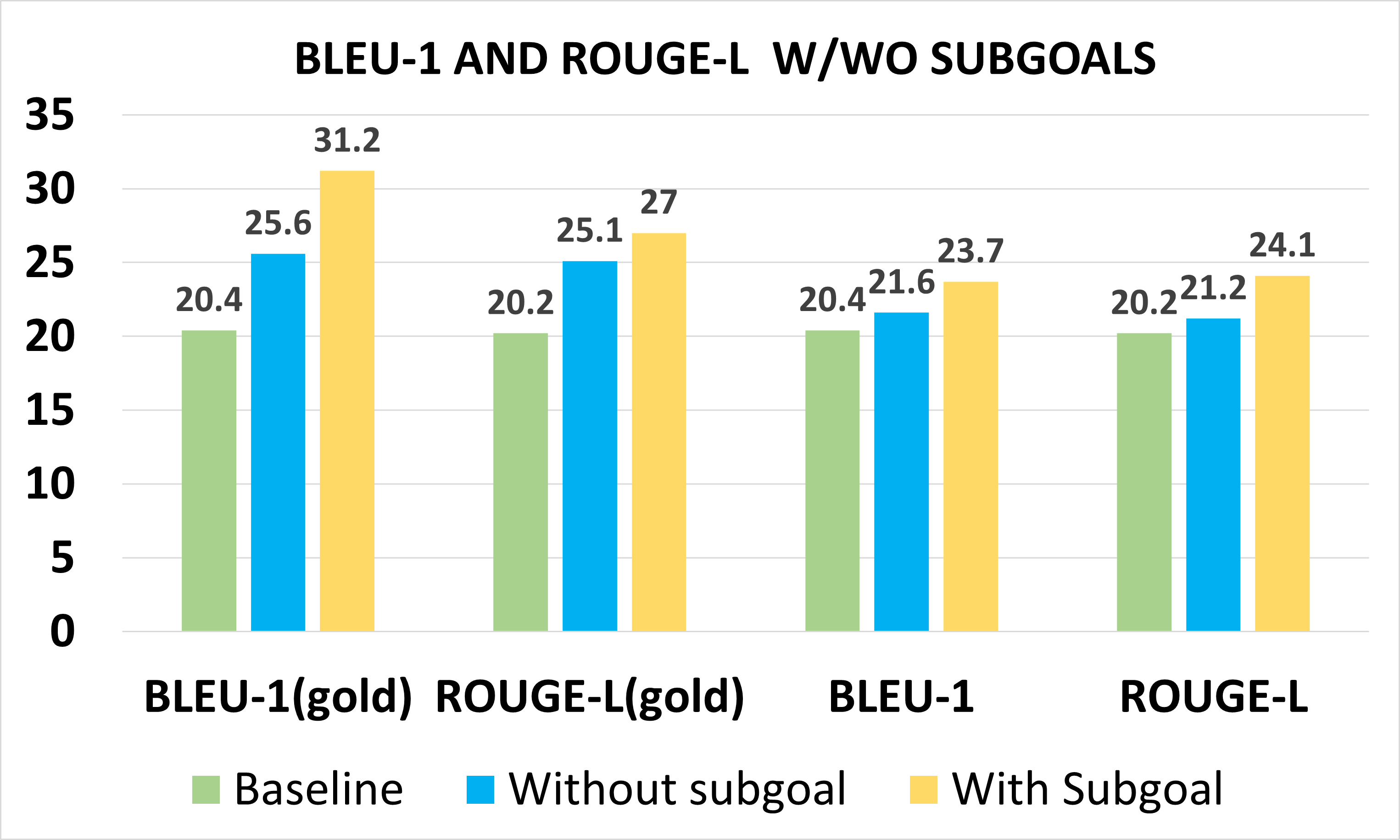}
\caption{BLEU-1 and ROUGE-L scores for generated scripts. Green color represents the baseline method. Blue color represents our method using segmentation separated with special tokens. Yellow color represents our method using both segmentation and subgoals.}\label{fig:goal_ablation2}
%\vspace{-12pt}
\end{figure}

\paragraph{Subgoals: Are these Necessary}
Since the automatic evaluation showcases an improvement in the quality of generated scripts, we hereby further investigate if such improvement is caused by the segmentation only. We experiment by formatting the training data scripts with special token at position of a new segment, but not adding the subgoals. We explored two settings. The first setting uses gold segmentation, and the second setting uses the predicted segmentation. We evaluated the scripts using the metrics in \cref{Experiment Settings}. The result in Figure~\ref{fig:goal_ablation2} shows that under both settings, simply with a few special tokens separating the steps in training dataset, the generated text improves in quality. However, the results are still worse than those with subgoals, explaining the significance of including subgoals in training data. In addition, for human, subgoals provide explainability for why models choose to make specific segments during script generation.

%YXCAO: 我觉得pipeline的方式，两个缺点，1是你说的这个，more repetition，因为缺少训练数据的原因。现实标注也很昂贵。2. 没有很好的利用结构信息去缩短dependence。比如人们写作也是一个段落一个段落，每个段落首句为中心句
\section{Case Study and Error Analysis}\label{case study1}
We present two example scripts generated using our method with the wikiHow dataset. Through these two examples, We analyze the common errors and the typical mistakes encountered. We put two more example scripts from Instructables dataset in \cref{case study} since they are rather long. These Instructables scripts manifest not only the common errors in this section but also a third typical mistake.

\begin{figure}[t]
\includegraphics[width=\linewidth]{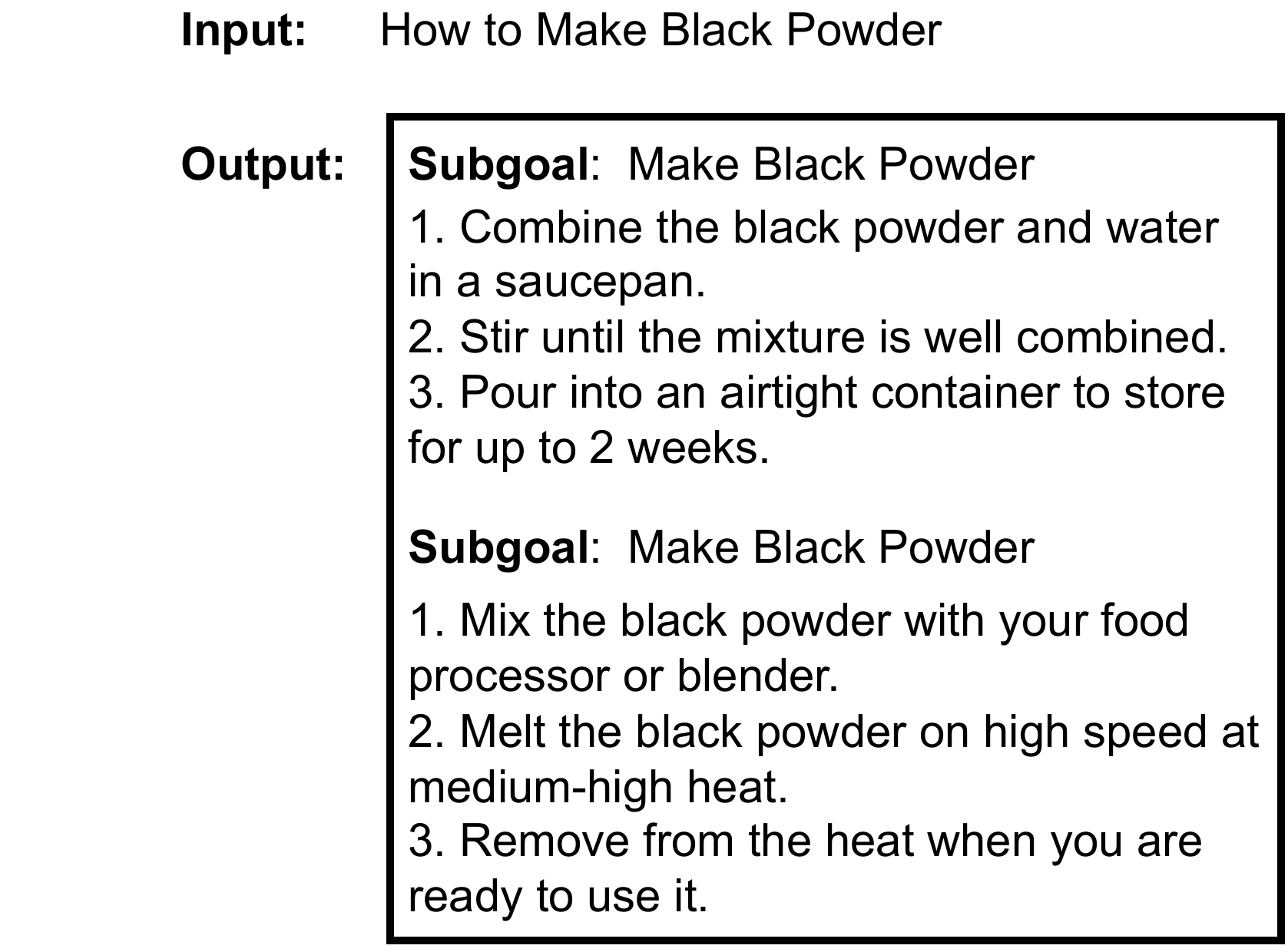}
\caption{The ``How to make black powder'' script (from wikiHow)}\label{fig:How to Make Black Powder}
\end{figure}

\begin{figure}[t]
\includegraphics[width=\linewidth]{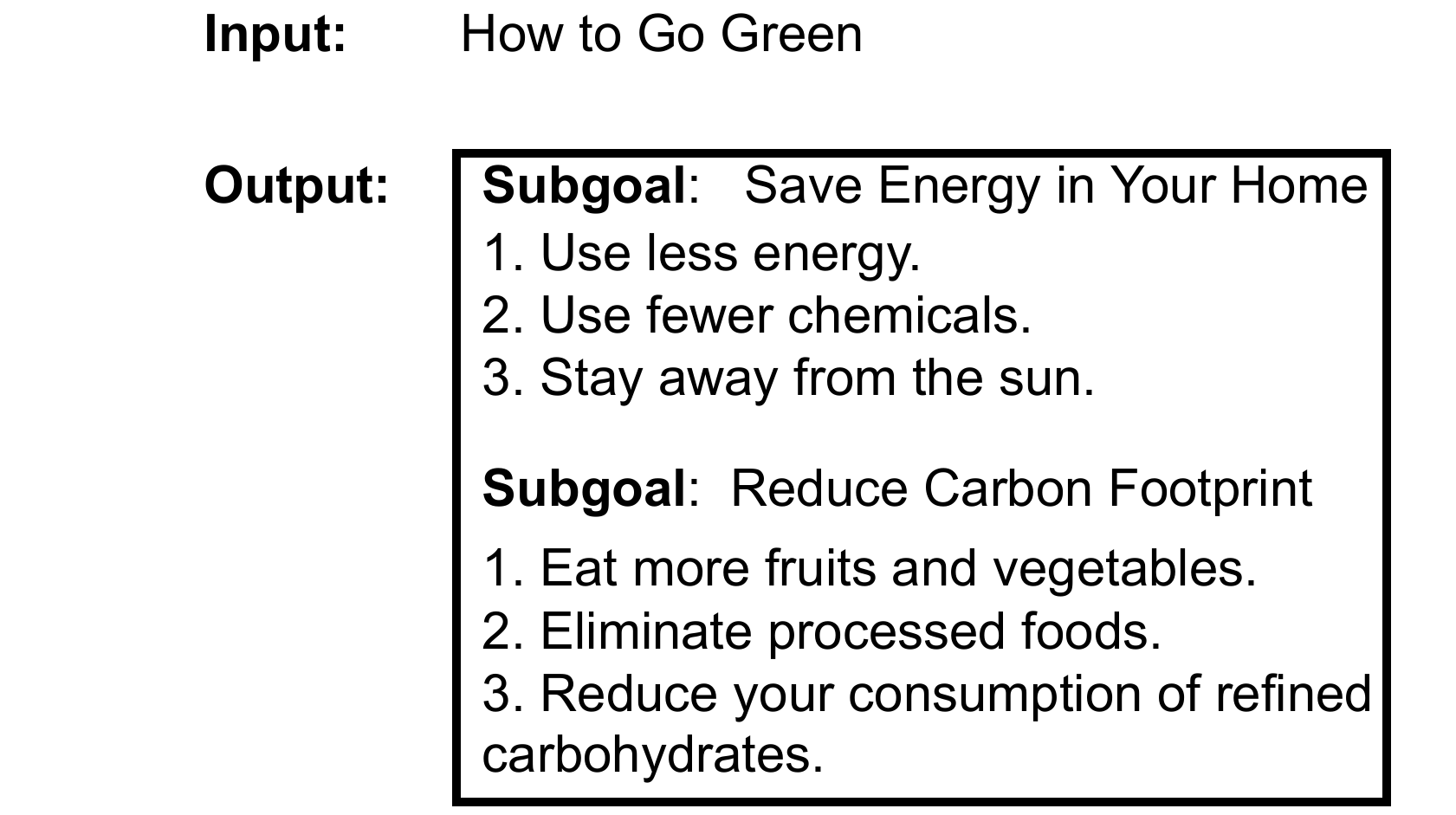}
\caption{The ``How to go green'' script (from wikiHow)}\label{fig:How to go green}
\end{figure}

\paragraph{Repetitive Subgoals}
According to our observation, the pervasive problem is that the generated subgoals are repeats of one another or the goal. This error appears in examples in Figure~\ref{fig:How to Make Black Powder}. One cause of this error is the inaccurate segmentation in the training dataset, which raises the difficulty of the subgoal prediction. Frequently, the subgoal labeled to a segment is no other but this segment's main goal. The reason is that the segment is a fraction of the script used as input when training the subgoal predictor. Moreover, the frequent occurrence of repetitive subgoals in the training dataset may seem like a pattern for the generation model, generating more scripts with repetitive subgoals. A revised loss function that penalizes repetition among goals and subgoals is a possible solution.

\paragraph{Irrelevant or Low-Quality Steps}
Another mistake with the generated scripts resides in the quality of the steps. For instance, in Figure~\ref{fig:How to Make Black Powder}, the model possibly mistakes ``black powder'' for ``black pepper'' and generates steps related to cooking. This mistake could originate from the lack of weapon-related knowledge in the training dataset. Figure~\ref{fig:How to go green} shows the script of ``go green,'' with two subgoals. Despite the reasonable subgoals generated, the steps under ``reduce carbon footprint'' are irrelevant. The correct interpretation of ``go green'' is about environmental-friendly measures, while the steps discuss ``green and healthy lifestyle.'' Since both wikiHow and Instructables use pictures to supplement text descriptions, a multi-modal approach may reduce ambiguity in the goal interpretation.
\section{Conclusion}
This work studies a new task of hierarchical script generation. To facilitate the research on this task, we contribute a new dataset Instructables to supplement the existing wikiHow resource, gathering over 100,000 hierarchical scripts. We further propose a method to build the benchmark, which learns to segment steps before script generation and concludes the subgoal each segment represents. Then, we fine-tune a T5-base model using prompts that combine subgoals and steps with special tokens. Experiment results from both automatic and human evaluation show that scripts generated by our method are in better quality than the baseline. Meanwhile, the gap towards the performance upper-bound still indicates much room for improvement. 
In the future, we are interested in exploring the idea of hierarchical generation in dealing with paragraphs in document generation.
%Future work will investigate the method to improve the generation quality further. Some potential areas for exploration include designing decoding algorithms to enhance generation conditioned on the subgoals.

\section*{Acknowledgement}

We appreciate the reviewers for their insightful
comments and suggestions.
Xinze, Yixin, and Aixin were supported under the RIE2020 Industry Alignment Fund – Industry Collaboration Projects (IAF-ICP) Funding Initiative, as well as cash and in-kind contribution from the industry partner(s). Yixin is supported by the Singapore Ministry of Education (MOE) Academic Research Fund (AcRF) Tier 1 grant. Muhao Chen was supported by the National Science Foundation of United States Grant IIS 2105329, an Amazon Research Award and a Cisco Research Award.

\section*{Limitation}
The dataset Intructables, same as wikiHow, has a two level hierarchy of goal-subgoals-steps. Such structure is because of the writing habits of human authors. Due to the lack of datasets with deeper nested hierarchy, our work does not investigate the cases when there are multiple levels of subgoals. In addition, this work focuses on the presentation of task and dataset, and does not explore the performance of more advanced language models on the task. Relevant studies can be conducted in future works.

\section*{Ethics Statement}
Our method is capable of generating large number of hierarchical scripts. Learning knowledge from different sources of information, the model might be misused into generating unsafe contents upon asking inappropriate questions. For now this seems unlikely since there is no offensive content in out collected dataset.

\bibliography{anthology,custom}

\begin{thebibliography}{33}
\expandafter\ifx\csname natexlab\endcsname\relax\def\natexlab#1{#1}\fi

\bibitem[{Antonietti et~al.(2000)Antonietti, Ignazi, and
  Perego}]{antonietti2000metacognitive}
Alessandro Antonietti, Sabrina Ignazi, and Patrizia Perego. 2000.
\newblock Metacognitive knowledge about problem-solving methods.
\newblock \emph{British Journal of Educational Psychology}, 70(1):1--16.

\bibitem[{Botvinick(2008)}]{botvinick2008hierarchical}
Matthew~M Botvinick. 2008.
\newblock Hierarchical models of behavior and prefrontal function.
\newblock \emph{Trends in cognitive sciences}, 12(5):201--208.

\bibitem[{Chaturvedi et~al.(2017)Chaturvedi, Peng, and
  Roth}]{chaturvedi-etal-2017-story}
Snigdha Chaturvedi, Haoruo Peng, and Dan Roth. 2017.
\newblock \href {https://doi.org/10.18653/v1/D17-1168} {Story comprehension for
  predicting what happens next}.
\newblock In \emph{Proceedings of the 2017 Conference on Empirical Methods in
  Natural Language Processing}, pages 1603--1614, Copenhagen, Denmark.
  Association for Computational Linguistics.

\bibitem[{Chen et~al.(2020)Chen, Zhang, Wang, and Roth}]{chen-etal-2020-trying}
Muhao Chen, Hongming Zhang, Haoyu Wang, and Dan Roth. 2020.
\newblock \href {https://doi.org/10.18653/v1/2020.conll-1.43} {What are you
  trying to do? semantic typing of event processes}.
\newblock In \emph{Proceedings of the 24th Conference on Computational Natural
  Language Learning}, pages 531--542, Online. Association for Computational
  Linguistics.

\bibitem[{Devlin et~al.(2019)Devlin, Chang, Lee, and
  Toutanova}]{devlin-etal-2019-bert}
Jacob Devlin, Ming-Wei Chang, Kenton Lee, and Kristina Toutanova. 2019.
\newblock \href {https://doi.org/10.18653/v1/N19-1423} {{BERT}: Pre-training of
  deep bidirectional transformers for language understanding}.
\newblock In \emph{Proceedings of the 2019 Conference of the North {A}merican
  Chapter of the Association for Computational Linguistics: Human Language
  Technologies, Volume 1 (Long and Short Papers)}, pages 4171--4186,
  Minneapolis, Minnesota. Association for Computational Linguistics.

\bibitem[{Fan et~al.(2018)Fan, Lewis, and Dauphin}]{fan-etal-2018-hierarchical}
Angela Fan, Mike Lewis, and Yann Dauphin. 2018.
\newblock \href {https://doi.org/10.18653/v1/P18-1082} {Hierarchical neural
  story generation}.
\newblock In \emph{Proceedings of the 56th Annual Meeting of the Association
  for Computational Linguistics (Volume 1: Long Papers)}, pages 889--898,
  Melbourne, Australia. Association for Computational Linguistics.

\bibitem[{Fan et~al.(2019)Fan, Lewis, and Dauphin}]{fan-etal-2019-strategies}
Angela Fan, Mike Lewis, and Yann Dauphin. 2019.
\newblock \href {https://doi.org/10.18653/v1/P19-1254} {Strategies for
  structuring story generation}.
\newblock In \emph{Proceedings of the 57th Annual Meeting of the Association
  for Computational Linguistics}, pages 2650--2660, Florence, Italy.
  Association for Computational Linguistics.

\bibitem[{Herman(1997)}]{herman_1997}
David Herman. 1997.
\newblock \href {https://doi.org/10.2307/463482} {Scripts, sequences, and
  stories: Elements of a postclassical narratology}.
\newblock \emph{PMLA/Publications of the Modern Language Association of
  America}, 112(5):1046–1059.

\bibitem[{Hu et~al.(2017)Hu, Yang, Liang, Salakhutdinov, and
  Xing}]{hu2017toward}
Zhiting Hu, Zichao Yang, Xiaodan Liang, Ruslan Salakhutdinov, and Eric~P Xing.
  2017.
\newblock Toward controlled generation of text.
\newblock In \emph{International conference on machine learning}, pages
  1587--1596. PMLR.

\bibitem[{Kingma and Ba(2014)}]{kingma2014adam}
Diederik~P Kingma and Jimmy Ba. 2014.
\newblock Adam: A method for stochastic optimization.
\newblock \emph{arXiv preprint arXiv:1412.6980}.

\bibitem[{Lagos et~al.(2017)Lagos, Gall{\'e}, Chernov, and
  S{\'a}ndor}]{lagos2017enriching}
Nikolaos Lagos, Matthias Gall{\'e}, Alexandr Chernov, and {\'A}gnes S{\'a}ndor.
  2017.
\newblock Enriching how-to guides with actionable phrases and linked data.
\newblock In \emph{Web Intelligence}, volume~15, pages 189--203. IOS Press.

\bibitem[{Lester et~al.(2021)Lester, Al-Rfou, and
  Constant}]{lester-etal-2021-power}
Brian Lester, Rami Al-Rfou, and Noah Constant. 2021.
\newblock \href {https://doi.org/10.18653/v1/2021.emnlp-main.243} {The power of
  scale for parameter-efficient prompt tuning}.
\newblock In \emph{Proceedings of the 2021 Conference on Empirical Methods in
  Natural Language Processing}, pages 3045--3059, Online and Punta Cana,
  Dominican Republic. Association for Computational Linguistics.

\bibitem[{Li et~al.(2016)Li, Galley, Brockett, Gao, and
  Dolan}]{li-etal-2016-diversity}
Jiwei Li, Michel Galley, Chris Brockett, Jianfeng Gao, and Bill Dolan. 2016.
\newblock \href {https://doi.org/10.18653/v1/N16-1014} {A diversity-promoting
  objective function for neural conversation models}.
\newblock In \emph{Proceedings of the 2016 Conference of the North {A}merican
  Chapter of the Association for Computational Linguistics: Human Language
  Technologies}, pages 110--119, San Diego, California. Association for
  Computational Linguistics.

\bibitem[{Lin(2004)}]{lin-2004-rouge}
Chin-Yew Lin. 2004.
\newblock \href {https://aclanthology.org/W04-1013} {{ROUGE}: A package for
  automatic evaluation of summaries}.
\newblock In \emph{Text Summarization Branches Out}, pages 74--81, Barcelona,
  Spain. Association for Computational Linguistics.

\bibitem[{Lu et~al.(2022)Lu, Welleck, West, Jiang, Kasai, Khashabi, Le~Bras,
  Qin, Yu, Zellers, Smith, and Choi}]{lu-etal-2022-neurologic}
Ximing Lu, Sean Welleck, Peter West, Liwei Jiang, Jungo Kasai, Daniel Khashabi,
  Ronan Le~Bras, Lianhui Qin, Youngjae Yu, Rowan Zellers, Noah~A. Smith, and
  Yejin Choi. 2022.
\newblock \href {https://doi.org/10.18653/v1/2022.naacl-main.57}
  {{N}euro{L}ogic a*esque decoding: Constrained text generation with lookahead
  heuristics}.
\newblock In \emph{Proceedings of the 2022 Conference of the North American
  Chapter of the Association for Computational Linguistics: Human Language
  Technologies}, pages 780--799, Seattle, United States. Association for
  Computational Linguistics.

\bibitem[{Lu et~al.(2021)Lu, West, Zellers, Le~Bras, Bhagavatula, and
  Choi}]{lu-etal-2021-neurologic}
Ximing Lu, Peter West, Rowan Zellers, Ronan Le~Bras, Chandra Bhagavatula, and
  Yejin Choi. 2021.
\newblock \href {https://doi.org/10.18653/v1/2021.naacl-main.339}
  {{N}euro{L}ogic decoding: (un)supervised neural text generation with
  predicate logic constraints}.
\newblock In \emph{Proceedings of the 2021 Conference of the North American
  Chapter of the Association for Computational Linguistics: Human Language
  Technologies}, pages 4288--4299, Online. Association for Computational
  Linguistics.

\bibitem[{Lyu et~al.(2021)Lyu, Zhang, and Callison-Burch}]{lyu-etal-2021-goal}
Qing Lyu, Li~Zhang, and Chris Callison-Burch. 2021.
\newblock \href {https://aclanthology.org/2021.inlg-1.19} {Goal-oriented script
  construction}.
\newblock In \emph{Proceedings of the 14th International Conference on Natural
  Language Generation}, pages 184--200, Aberdeen, Scotland, UK. Association for
  Computational Linguistics.

\bibitem[{Modi and Titov(2014)}]{modi-titov-2014-inducing}
Ashutosh Modi and Ivan Titov. 2014.
\newblock \href {https://doi.org/10.3115/v1/W14-1606} {Inducing neural models
  of script knowledge}.
\newblock In \emph{Proceedings of the Eighteenth Conference on Computational
  Natural Language Learning}, pages 49--57, Ann Arbor, Michigan. Association
  for Computational Linguistics.

\bibitem[{M{\"u}llner(2011)}]{mullner2011modern}
Daniel M{\"u}llner. 2011.
\newblock Modern hierarchical, agglomerative clustering algorithms.
\newblock \emph{arXiv preprint arXiv:1109.2378}.

\bibitem[{Papineni et~al.(2002)Papineni, Roukos, Ward, and
  Zhu}]{papineni-etal-2002-bleu}
Kishore Papineni, Salim Roukos, Todd Ward, and Wei-Jing Zhu. 2002.
\newblock \href {https://doi.org/10.3115/1073083.1073135} {{B}leu: a method for
  automatic evaluation of machine translation}.
\newblock In \emph{Proceedings of the 40th Annual Meeting of the Association
  for Computational Linguistics}, pages 311--318, Philadelphia, Pennsylvania,
  USA. Association for Computational Linguistics.

\bibitem[{Pareti et~al.(2014)Pareti, Testu, Ichise, Klein, and
  Barker}]{pareti2014integrating}
Paolo Pareti, Benoit Testu, Ryutaro Ichise, Ewan Klein, and Adam Barker. 2014.
\newblock Integrating know-how into the linked data cloud.
\newblock In \emph{International Conference on Knowledge Engineering and
  Knowledge Management}, pages 385--396. Springer.

\bibitem[{Peng et~al.(2021)Peng, Li, Li, Shayandeh, Liden, and
  Gao}]{peng-etal-2021-soloist}
Baolin Peng, Chunyuan Li, Jinchao Li, Shahin Shayandeh, Lars Liden, and
  Jianfeng Gao. 2021.
\newblock \href {https://doi.org/10.1162/tacl_a_00399} {Soloist: Building task
  bots at scale with transfer learning and machine teaching}.
\newblock \emph{Transactions of the Association for Computational Linguistics},
  9:807--824.

\bibitem[{Pichotta and Mooney(2016)}]{pichotta-mooney-2016-using}
Karl Pichotta and Raymond~J. Mooney. 2016.
\newblock \href {https://doi.org/10.18653/v1/P16-1027} {Using sentence-level
  {LSTM} language models for script inference}.
\newblock In \emph{Proceedings of the 54th Annual Meeting of the Association
  for Computational Linguistics (Volume 1: Long Papers)}, pages 279--289,
  Berlin, Germany. Association for Computational Linguistics.

\bibitem[{Raffel et~al.(2020)Raffel, Shazeer, Roberts, Lee, Narang, Matena,
  Zhou, Li, Liu et~al.}]{raffel2020exploring}
Colin Raffel, Noam Shazeer, Adam Roberts, Katherine Lee, Sharan Narang, Michael
  Matena, Yanqi Zhou, Wei Li, Peter~J Liu, et~al. 2020.
\newblock Exploring the limits of transfer learning with a unified text-to-text
  transformer.
\newblock \emph{J. Mach. Learn. Res.}, 21(140):1--67.

\bibitem[{Reimers and Gurevych(2019)}]{reimers-2019-sentence-bert}
Nils Reimers and Iryna Gurevych. 2019.
\newblock \href {https://arxiv.org/abs/1908.10084} {Sentence-bert: Sentence
  embeddings using siamese bert-networks}.
\newblock In \emph{Proceedings of the 2019 Conference on Empirical Methods in
  Natural Language Processing}. Association for Computational Linguistics.

\bibitem[{Wang et~al.(2022)Wang, Xu, Szekely, and Chen}]{wang2022robust}
Fei Wang, Zhewei Xu, Pedro Szekely, and Muhao Chen. 2022.
\newblock Robust (controlled) table-to-text generation with structure-aware
  equivariance learning.
\newblock \emph{arXiv preprint arXiv:2205.03972}.

\bibitem[{Yao et~al.(2019)Yao, Mao, and Luo}]{yao2019kg}
Liang Yao, Chengsheng Mao, and Yuan Luo. 2019.
\newblock Kg-bert: Bert for knowledge graph completion.
\newblock \emph{arXiv preprint arXiv:1909.03193}.

\bibitem[{Zhang et~al.(2020{\natexlab{a}})Zhang, Chen, Wang, Song, and
  Roth}]{zhang2020analogous}
Hongming Zhang, Muhao Chen, Haoyu Wang, Yangqiu Song, and Dan Roth.
  2020{\natexlab{a}}.
\newblock Analogous process structure induction for sub-event sequence
  prediction.
\newblock \emph{arXiv preprint arXiv:2010.08525}.

\bibitem[{Zhang and Norman(1994)}]{zhang1994representations}
Jiaje Zhang and Donald~A Norman. 1994.
\newblock Representations in distributed cognitive tasks.
\newblock \emph{Cognitive science}, 18(1):87--122.

\bibitem[{Zhang et~al.(2020{\natexlab{b}})Zhang, Lyu, and
  Callison-Burch}]{zhang-etal-2020-reasoning}
Li~Zhang, Qing Lyu, and Chris Callison-Burch. 2020{\natexlab{b}}.
\newblock \href {https://doi.org/10.18653/v1/2020.emnlp-main.374} {Reasoning
  about goals, steps, and temporal ordering with {W}iki{H}ow}.
\newblock In \emph{Proceedings of the 2020 Conference on Empirical Methods in
  Natural Language Processing (EMNLP)}, pages 4630--4639, Online. Association
  for Computational Linguistics.

\bibitem[{Zhang et~al.(2020{\natexlab{c}})Zhang, Chen, and
  Bui}]{zhang2020diagnostic}
Tianran Zhang, Muhao Chen, and Alex~AT Bui. 2020{\natexlab{c}}.
\newblock Diagnostic prediction with sequence-of-sets representation learning
  for clinical events.
\newblock In \emph{International Conference on Artificial Intelligence in
  Medicine}, pages 348--358. Springer.

\bibitem[{Zhang et~al.(2019)Zhang, Kishore, Wu, Weinberger, and
  Artzi}]{zhang2019bertscore}
Tianyi Zhang, Varsha Kishore, Felix Wu, Kilian~Q Weinberger, and Yoav Artzi.
  2019.
\newblock Bertscore: Evaluating text generation with bert.
\newblock \emph{arXiv preprint arXiv:1904.09675}.

\bibitem[{Zhou et~al.(2022)Zhou, Zhang, Yang, Lyu, Yin, Callison-Burch, and
  Neubig}]{zhou-etal-2022-show}
Shuyan Zhou, Li~Zhang, Yue Yang, Qing Lyu, Pengcheng Yin, Chris Callison-Burch,
  and Graham Neubig. 2022.
\newblock \href {https://doi.org/10.18653/v1/2022.acl-long.214} {Show me more
  details: Discovering hierarchies of procedures from semi-structured web
  data}.
\newblock In \emph{Proceedings of the 60th Annual Meeting of the Association
  for Computational Linguistics (Volume 1: Long Papers)}, pages 2998--3012,
  Dublin, Ireland. Association for Computational Linguistics.

\end{thebibliography}
\bibliographystyle{acl_natbib}
\appendix
\newpage

\section{Comparison between wikiHow and Instructables} \label{app: compare}
In this section we compare the features and statistics between dataset Instructables and wikiHow in detail. 

The two datasets are different in multiple aspects. In terms of content, Instructables includes innovative thoughts on building concrete objects (e.g., \textit{toy rocket}), While wikiHow incorporates daily-life experiences for possibly abstract concepts (e.g., \textit{live healthily}). In terms of language style, Instructables is subjective (e.g., \textit{I built it with ...}), and wikiHow is relatively objective with the use of imperatives (e.g., \textit{Take a good sleep}). Regarding domain, Instructables involves six domains like circuits and craft, while wikiHow spans over 19 domains like arts and sports.

For Instructables dataset, on average, there are 5.2 subgoals for each script, 2.6 steps per subgoal, and 18.1 words per step. We also collect this statistics for wikiHow dataset, which includes 112,451 scripts, 278,680 subgoals, 2,057,088 steps, and 12,281,074 words. For wikiHow, there are 2.5 subgoals for each script, 7.4 steps per subgoal, and 6.0 words per step on average. The average sentence length of Instructables is much longer due to its narrative-based language style. Compared to wikiHow, which focuses on daily-life experiences, Instructables is more challenging since many items to build are highly professional and complicated (i.e., a Blind Assist Ultrasonic Navigator), which also explains the reason for the large average number of subgoals per script.

\section{Segmentation Methods}\label{app: segmentation}
In this section, we formally explain the algorithms and implementations of each segmentation method in detail.

\subsection{Next Sentence Prediction}\label{app: NSP}
We separate two consecutive steps if their continuity is predicted as negative via next sentence prediction --- the two steps are talking about different topics. Given a list of ordered steps, we concatenate every two consecutive steps as \textit{[CLS]step1[SEP]step2[SEP]} and calculate the probability score using BERT-base \cite{devlin-etal-2019-bert} model. Specifically, we assume that a higher probability score indicates that the latter step is more rational than the previous one. We determine $K$ lowest probability scores corresponding to $K$ segmentation points. In experiments, we heuristically find the best $K$ between 2 to 3.

\subsection{Perplexity}\label{app: perplexity}
Another approach is to measure the plausibility of a list of steps with \textbf{perplexity}. Assume that for a list of steps [$S_x$ to $S_y$], the gold segment position is between $S_i$ and $S_{i+1}$ ($x$ < $i$ < $y$), separating the list into two segments [$S_x$ to $S_i$] and [$S_{i+1}$ to $S_y$]. The perplexity of [$S_x$ to $S_{i+1}$] should be greater than that of [$S_x$ to $S_i$] since an additional sentence not belonging to the segment makes it less natural. Similarly, the perplexity of [$S_i$ to $S_y$] should be greater than that of [$S_{i+1}$ to $S_y$]. We iterate from $i$ = $0$ to $i$ = $N-1$ (number of steps), and mark the $i$ as a segmentation point if it satisfies both perplexity requirements.

\subsection{Agglomerative Clustering}\label{app: agglomerative clustering}
Instead of looking for segmentation points, we apply hierarchical agglomerative clustering (HAC) \cite{mullner2011modern} to group steps based on their sentence embeddings using SentenceBert~\cite{reimers-2019-sentence-bert}. Specifically, we merge two steps if their euclidean distance falls below a threshold while maintaining variance within all clusters minimized. Since HAC does not guarantee consecutive steps in the same cluster (e.g., if steps 1,2,4,5 are in cluster A, step 3 could be in step B), we make adjustment by recursively sending each step to the cluster with most of its  neighbours, and sort the steps in the end.

\begin{algorithm}[t]
{
\small
\caption{Finding segmentation points with topic detecting. N is the number of steps in the script. x and y are start and end positions of subset. S is the list of segmentation points we look for.}\label{alg:fast}
\begin{algorithmic}
\Require $N \geq 3$ \Comment{At least 3 steps in a script}
\State $x \gets 0$
\State $y \gets 2$
\State $S \gets list$
\While{$y < N$}
\If{$topicNumber(x,y) < 2$}
    \State $y \gets y + 1$
\ElsIf{$topicNumber(x,y) \geq 2$}
    \State $S \gets y - 2$
    \State $x \gets y - 1$
    \State $y \gets y + 1$
\EndIf
\EndWhile
\end{algorithmic}
}
\end{algorithm}

\subsection{Topic Detecting}\label{app: topic detecting}
While NSP compares topics locally between 2 steps, we design this method to detect topics globally among multiple steps. As shown in algorithm~\ref{alg:fast}, starting with the first two steps, we add one step each time. A segmentation point is marked before the new step if more than one topic is detected. The topic detecting is implemented using fastclustering\footnote{\url{https://www.sbert.net/examples/applications/clustering/README.html}}, which calculates cosine-similarity scores among the steps based on their sentence embeddings. Assuming that steps that share a topic have higher similarity scores, steps are assigned to the same community if their scores are above a threshold. In practice, we find 0.65 a reasonable threshold.

\section{Model Configuration}\label{app: model configuration}
We fine-tune the T5 model from the Hugging-Face service\footnote{\url{https://huggingface.co/t5-base}}. 
We use Adam \cite{kingma2014adam} for optimization with the learning rate of 1e-4. We set the batch size 16 to fit the memory of one NVIDIA Tesla v100 GPU. 
%The experiment takes 6 hours to 12 hours in the worst case. 
The number of epochs is limited to 3 for models to converge within a reasonable running time. 
Training takes around 6 hours to finish on wikiHow and 12 hours on Instructables.
We choose the model with the highest development performance to be evaluated on the test set.

\section{Human Evaluation Details}\label{app: human evaluation questions}
\begin{figure}[ht]
\includegraphics[width=\linewidth]{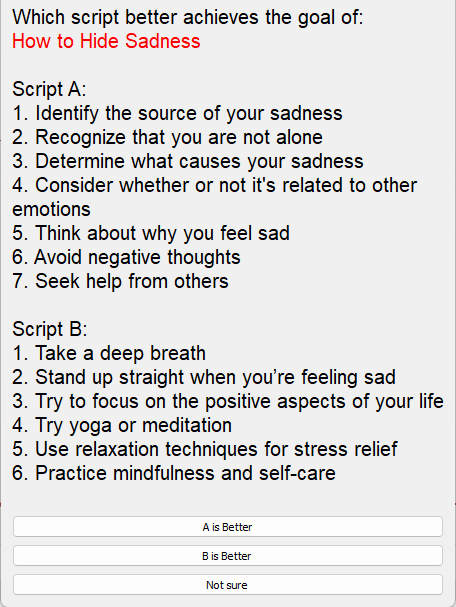}
\caption{Screenshot of a question shown to the annotators, asking them to select the script that achieves the goal better from the two.}\label{fig:human_screenshot_steps}
\end{figure}

\begin{figure}[ht]
\includegraphics[width=\linewidth]{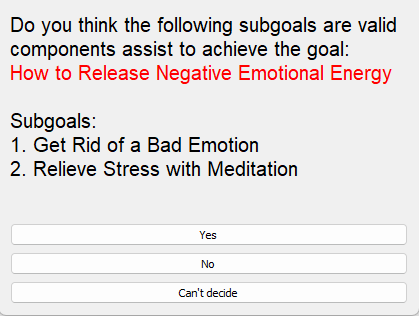}
\caption{Screenshot of a question shown to the annotators, asking them to judge if the subgoals generated are valid components of the given goal, and are helpful in achieving the goal.}\label{fig:human_screenshot_goal_subgoal}
\end{figure}

\begin{figure}[ht]
\includegraphics[width=\linewidth]{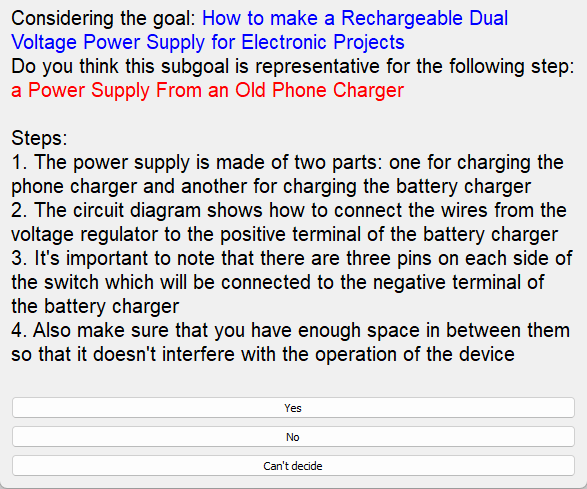}
\caption{Screenshot of a question shown to the annotators, asking them to judge if the generated subgoals are representative of the steps, considering the provided goal.}\label{fig:human_screenshot_subgoal_step}
\end{figure}
In this section, we explain in detail our human evaluation settings. For step evaluation, each question provides a goal and two scripts, asking the annotator which script better achieves the goal. Three options are provided, ``A is better'', ``B is better'', and ``Not sure''. Note that we flatten the script generated by our method and randomize the positions (A or B) of scripts in different questions to prevent the annotators from possibly identifying which script is ours from non-content information. For subgoal evaluation, we evaluate the association between 1) goal and subgoal and 2) subgoal and step. For goal-subgoal evaluation, each question provides a goal and a list of subgoals generated, asking the annotators if the subgoals are valid components of the goal and assist in achieving the goal. For subgoal-step evaluation, each question provides a goal, a subgoal, and a list of steps, asking the annotators if the subgoal is representative of the steps considering the goal.Three options are provided, ``Yes'', ``No'', and ``Can't decide''.

For each criterion (step, goal-subgoal, and subgoal-step), we randomly generate 100 questions from the test set of each dataset (wikiHow, Instructables), giving a total of 600 questions. We employ four human annotators in total. All annotators are graduate students and native or proficient speakers of English. All annotators possess adequate real-life knowledge and experiences to make reasonable judgments about the provided goals and have no potential conflicts of interest in this work. Each set of questions is answered by two different annotators, and for any disagreement, a third annotator will provide the final answer.

We hereby present the screenshots of the human evaluation questions. Figure~\ref{fig:human_screenshot_steps} corresponds to the questions which compare the script generated by our method with the baseline. Figure~\ref{fig:human_screenshot_goal_subgoal} and \ref{fig:human_screenshot_subgoal_step} correspond to the questions which evaluate the quality of generated subgoals

Since each question is answered by two annotators, we report Inter Annotator Agreement (IAA) by providing the number of questions (out of 100) that a third annotator is not required. For the question "Which script better achieves the goal", the IAA are 72 for wikiHow and 81 for Instructables. For the question "Are the subgoals helpful to achieve the goal", the IAA are 78 for wikiHow and 50 for Instructables. For the question "Is the subgoal representative of the steps", the IAA are 69 for wikiHow and 70 for Instructables.

%\section{Human Evaluation Results} \label{app: human evaluation results}
%In this section, we show the results of human evaluation in stacked bar charts. Note that for the question where we ask annotators to select the better script (Figure~\ref{fig:human_eval_steps}), we allow ``not sure'' to be the final result after coordination since it is a comparison question. It is possible that both scripts accomplish the given goal equally well or neither script accomplishes the goal. For the two questions evaluating the subgoals (Figure~\ref{fig:human_eval_goal_subgoal} and \ref{fig:human_eval_subgoal_step}), all answers have to be solid ``Yes'' or ``No'' after coordination.

\section{More Case Studies}\label{case study}
In this section we further analyze two hierarchical scripts generated using our method from Instructables dataset.

\begin{figure}[ht]
\includegraphics[width=\linewidth]{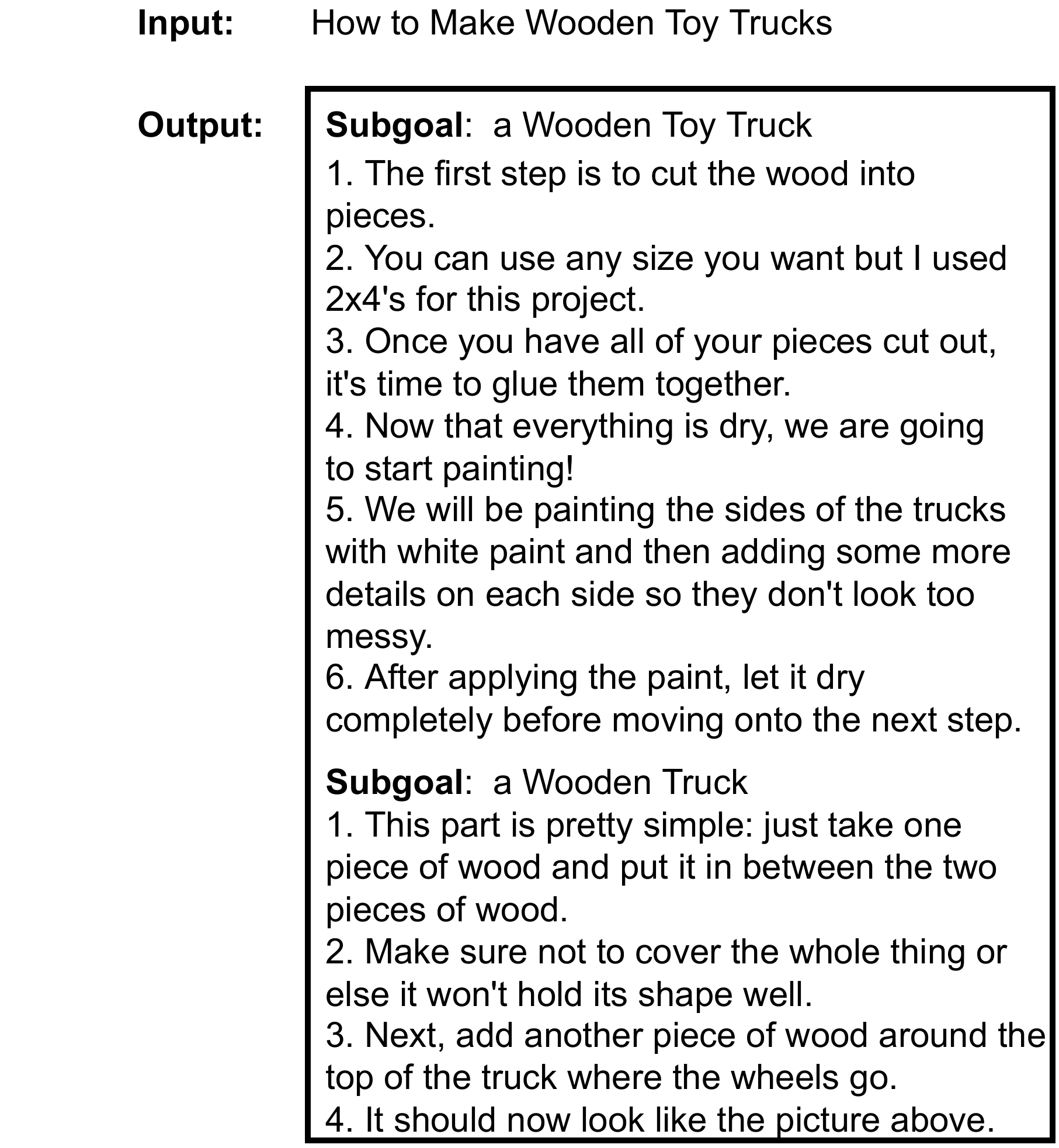}
\caption{The ``How to make wooden toy trucks'' script (from Instructables)}\label{fig:How to make wooden toy trucks}
\end{figure}

The example in Figure~\ref{fig:How to make wooden toy trucks} also manifests the two common errors presented in \cref{case study1}, \textbf{repetitive subgoals} and \textbf{low-quality steps}. To elaborate,  the steps ambiguously mention the concepts of processing the woods, painting, and assembling the wheel and truck body. However, the overall quality of the steps are rather low and does not lead to the fulfillment of the goal. A goal like this with complicated building steps may still be challenging for models at T5-base level.

\begin{figure}[ht]
\includegraphics[width=\linewidth]{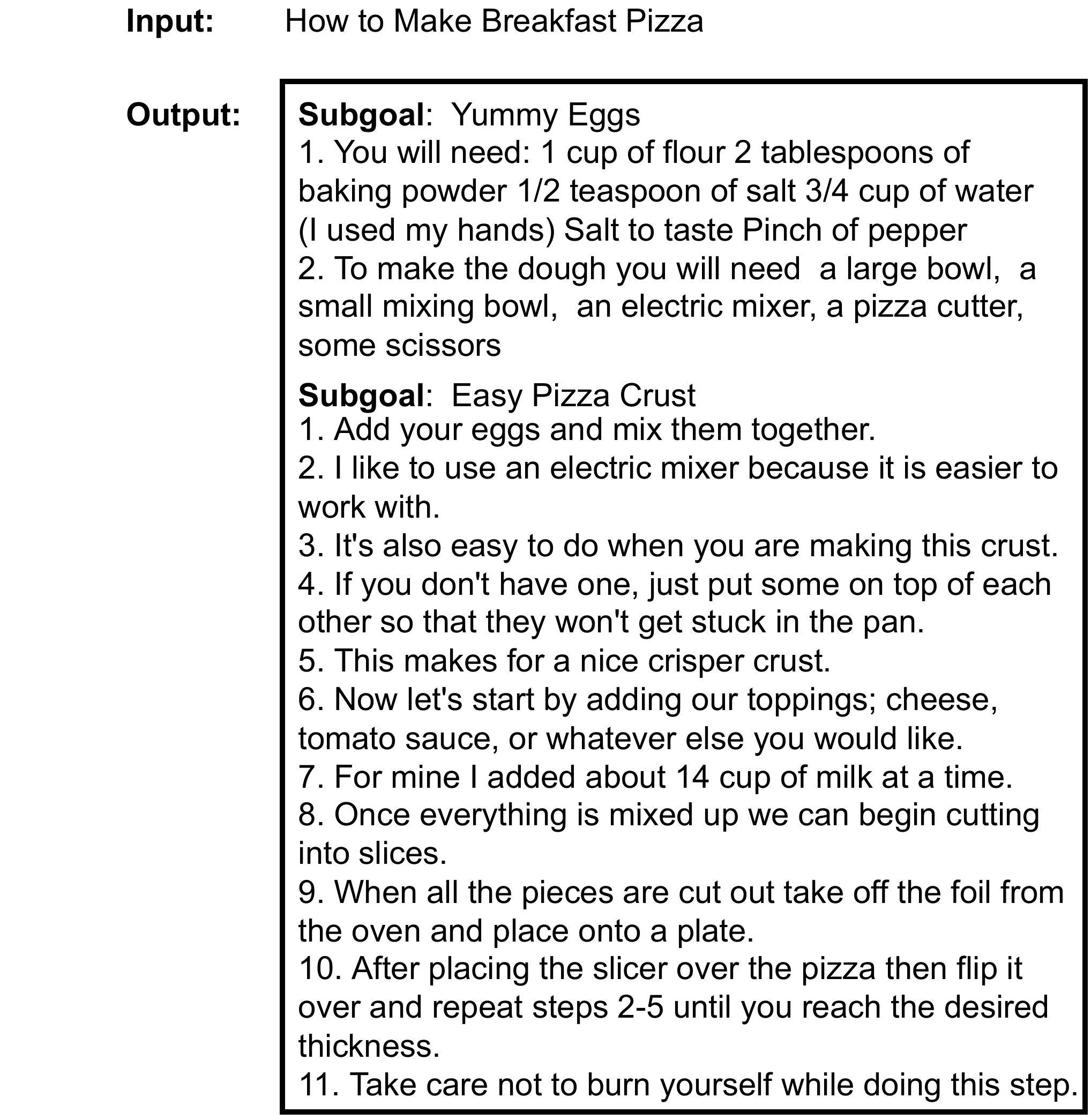}
\caption{The ``How to make breakfast pizza'' script (from Instructables)}\label{fig:How to make breakfast pizza}
\end{figure}

Another problem observed is \textbf{inaccurate segmentation or subgoal}. Each subgoal is supposed to be a good summary of the corresponding steps. The example in Figure~\ref{fig:How to make breakfast pizza} shows an incorrect subgoal, ``Yummy Eggs'', while the steps are about preparation. The ``egg'' possibly comes from the keyword ``breakfast'' at decoding time. The second subgoal is not perfect either, whereby the last three steps discuss the idea of ``baking'' instead of ``making pizza crust''; hence they should be separated into a third segment. This error manifests as an area of improvement in models' ability to understand the big picture of the event. It might be helpful to revise the decoding strategy for beam search in the future. By adding heuristics~\cite{lu-etal-2022-neurologic} in beam search, it is possible to improve the constrained generation process to look into the goal globally instead of locally.

\section{More Qualitative Examples}
\begin{figure*}[ht]
\includegraphics[width=\linewidth]{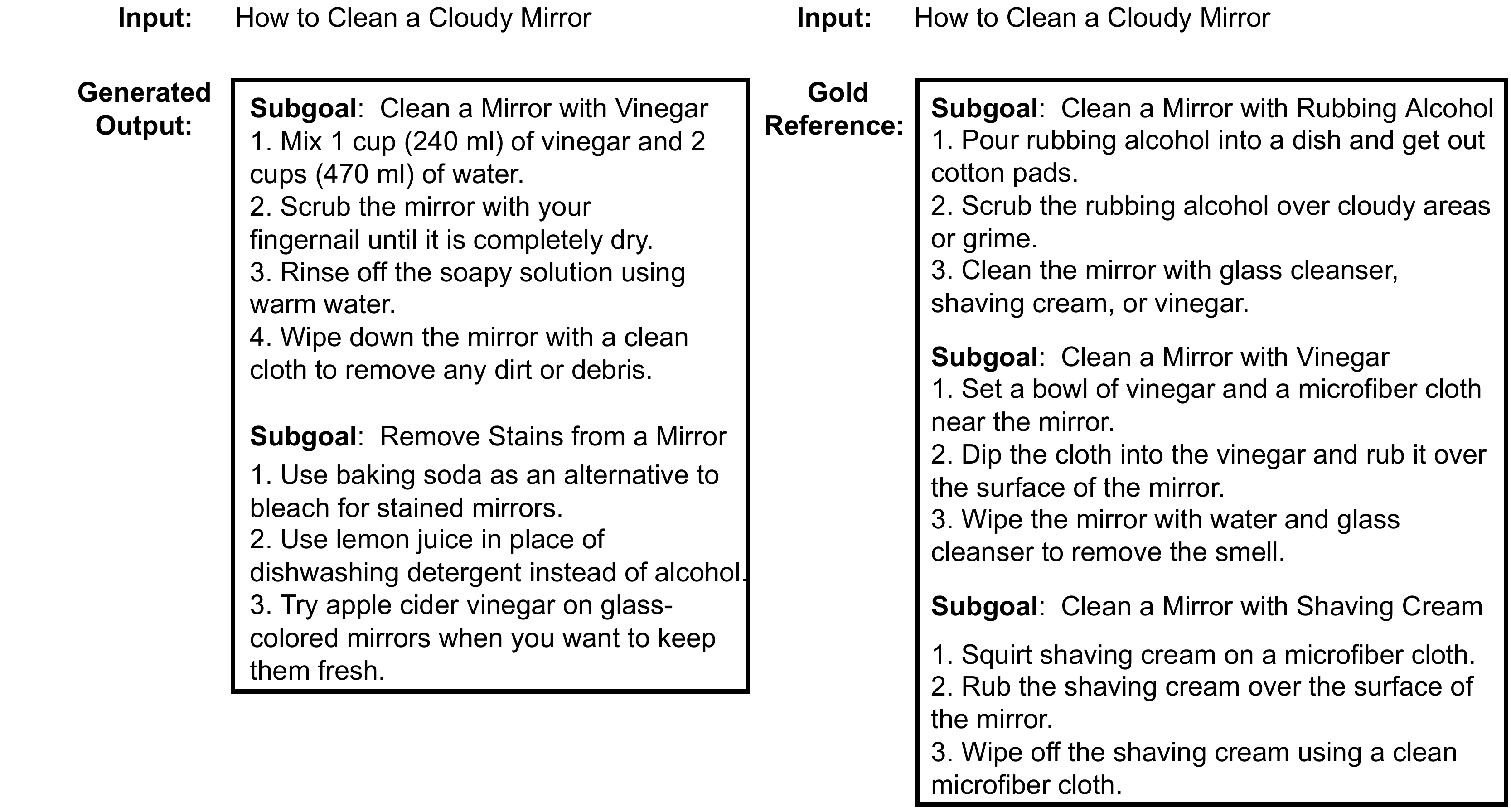}
\caption{The ``How to clean a cloudy mirror'' script (from wikiHow)}\label{fig:clean a cloudy mirror}
\end{figure*}

\begin{figure*}[ht]
\includegraphics[width=\linewidth]{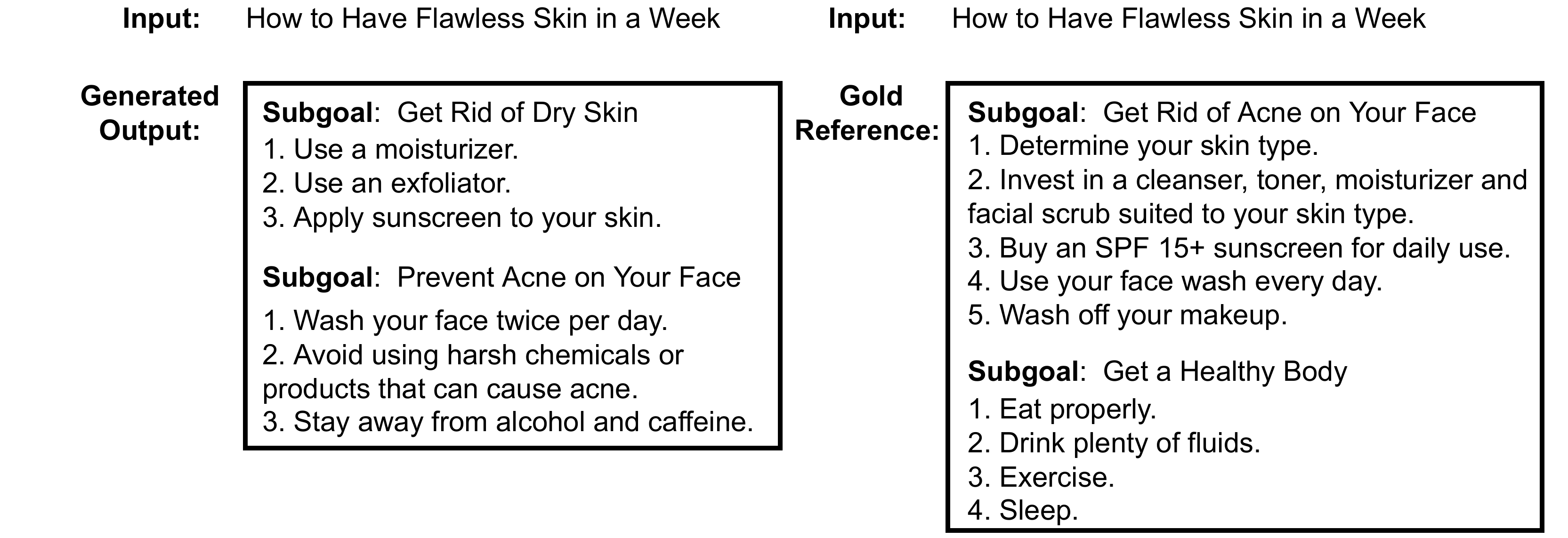}
\caption{The ``How to have flawless skin in a Week'' script (from wikiHow)}\label{fig:have flawless skin}
\end{figure*}

\begin{figure*}[ht]
\includegraphics[width=\linewidth]{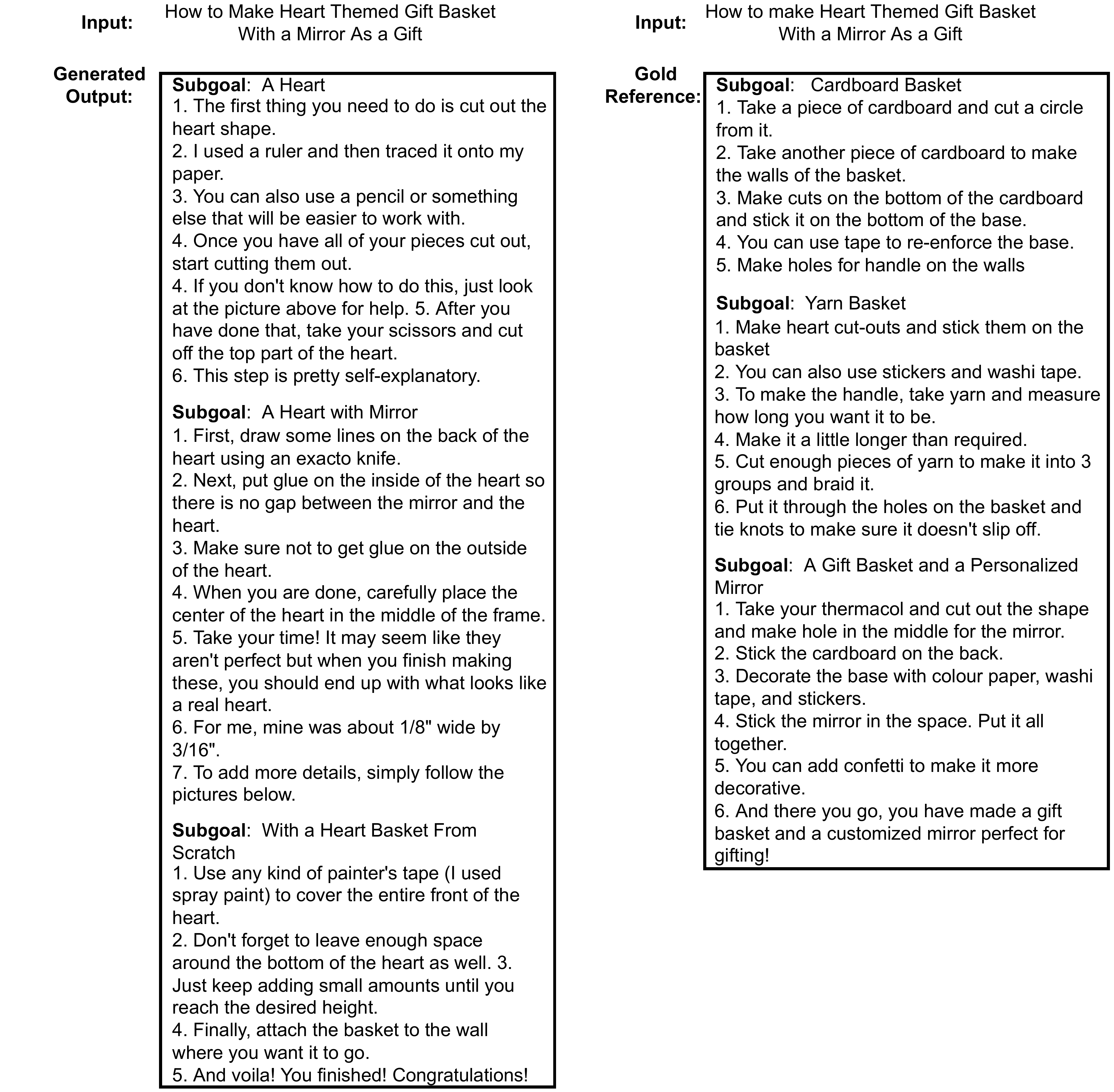}
\caption{The ``How to make heart themed gift basket with a mirror as a gift'' script (from Instructables)}\label{fig:heart themed gift basket}
\end{figure*}

Aside from the examples shown in section \cref{case study},  We provide 3 examples scripts generated and compare them with the ground truth. The outputs on the left hand side are generated with our method and the scripts on the right hand side are the gold references composed by human authors on wikiHow and Instructables websites respectively. Please see Figure~\ref{fig:clean a cloudy mirror} and \ref{fig:have flawless skin} for wikiHow, and Figure~\ref{fig:heart themed gift basket} for Instructables. 

\end{document}